\documentclass{ecai} 
\usepackage{microtype}
\usepackage{graphicx}
\usepackage{booktabs} % for professional tables
\usepackage{multirow} % In case you need multi-row cells in tables            % For including images
\usepackage{subcaption} 
\usepackage{mathtools}
\usepackage{float}
\usepackage{bm}

\usepackage{latexsym}
\usepackage{amssymb}
\usepackage{amsmath}
\usepackage{amsthm}
\usepackage{booktabs}
\usepackage{enumitem}
\usepackage{graphicx}
\usepackage{color}
\usepackage{hyperref}
\usepackage{algorithm}
\usepackage{algorithmic}

\newcommand{\BibTeX}{B\kern-.05em{\sc i\kern-.025em b}\kern-.08em\TeX}

\begin{document}

%%%%%%%%%%%%%%%%%%%%%%%%%%%%%%%%%%%%%%%%%%%%%%%%%%%%%%%%%%%%%%%%%%%%%%%%

\begin{frontmatter}

%%% Use this command to specify your submission number.
%%% In doubleblind mode, it will be printed on the first page.

\paperid{6425} 

%%% Use this command to specify the title of your paper.

\title{Contrast All The Time: Learning Time Series Representation from Temporal Consistency}

%%% Use this combinations of commands to specify all authors of your 
%%% paper. Use \fnms{} and \snm{} to indicate everyone's first names 
%%% and surname. This will help the publisher with indexing the 
%%% proceedings. Please use a reasonable approximation in case your 
%%% name does not neatly split into "first names" and "surname".
%%% Specifying your ORCID digital identifier is optional. 
%%% Use the \thanks{} command to indicate one or more corresponding 
%%% authors and their email address(es). If so desired, you can specify
%%% author contributions using the \footnote{} command.

\author[A]{\fnms{Abdul-Kazeem}~\snm{Shamba}\thanks{Corresponding Author. Email: abdul.k.shamba@ntnu.no.}}
\author[A]{\fnms{Kerstin}~\snm{Bach}}
\author[B]{\fnms{Gavin}~\snm{Taylor}}

\address[A]{Department of Computer Science, Norwegian University of Science and Technology, Norway}
\address[C]{Department of Computer Science, United States Naval Academy, USA}

%%% Use this environment to include an abstract of your paper.

\begin{abstract}
Representation learning for time series using contrastive learning has emerged as a critical technique for improving the performance of downstream tasks. To advance this effective approach, we introduce CaTT (\textit{Contrast All The Time}), a new approach to unsupervised contrastive learning for time series, which takes advantage of dynamics between temporally similar moments more efficiently and effectively than existing methods. CaTT departs from conventional time-series contrastive approaches that rely on data augmentations or selected views. Instead, it uses the full temporal dimension by contrasting all time steps in parallel. This is made possible by a scalable NT-pair formulation, which extends the classic N-pair loss across both batch and temporal dimensions, making the learning process end-to-end and more efficient. CaTT learns directly from the natural structure of temporal data, using repeated or adjacent time steps as implicit supervision, without the need for pair selection heuristics. We demonstrate that this approach produces superior embeddings which allow better performance in downstream tasks. Additionally, training is faster than other contrastive learning approaches, making it suitable for large-scale and real-world time series applications. The source code is publicly available at \href{https://github.com/sfi-norwai/CaTT}{https://github.com/sfi-norwai/CaTT}.
\end{abstract}

\end{frontmatter}

%%%%%%%%%%%%%%%%%%%%%%%%%%%%%%%%%%%%%%%%%%%%%%%%%%%%%%%%%%%%%%%%%%%%%%%%

\section{Introduction}

Much of the world's data, including time series (TS) data, lies in large, unlabeled datasets. Working with these datasets naively requires extensive data annotation, costing money, effort, and time. Self-supervised learning, which produces descriptive and intelligible representations in natural language processing (NLP) and computer vision (CV) \citep{cpc, simclr, dino, dinov2}, has emerged as a promising path for learning time series (TS) representations, which allow for coherent data organization before any labeling or annotation, easing the time and domain expertise required for understanding these datasets. One approach to representation learning is contrastive learning, in which similar (positive) and dissimilar (negative) pairs of samples are identified, and the embeddings of positive pairs are adjusted to be closer, and the embeddings of negative pairs are adjusted to be further. For this to be effective, positive and negative pairs must be identified in an unsupervised way accurately and efficiently, and training must be quick. In CV, data augmentation has been successful in creating positive pairs in an unsupervised way; however, in TS analysis, this can introduce inductive bias \citep{ts2vec, infots, cpc, tnc, cost}.  In TS analysis, it is instead possible to use temporal context, and create positive pairs on the assumption that moments close in time should have similar embeddings. However, TS representation learning approaches that leverage temporal information in the contrastive objective struggle with scalability and speed. These limitations arise from inefficient positive pair identification and usage, where only a subset of the training batch is used for the contrastive objective. As a result, these approaches are poorly suited for real-world applications, particularly in dynamic streaming data environments.

To address these challenges, we revisit hard contrastive learning and propose a novel framework for unsupervised contrastive representation learning in time series, \textit{contrast all the time} (CaTT). CaTT adapts an N-pair loss introduced by \citet{npair} to use {\it every} sample in a batch as a positive or negative pair without an additional identification step, making the learning process end-to-end and more efficient. The N-pair loss solves the problem of selecting statistically relevant instances in every batch \citep{tnc}. Inspired by the finite difference heat equation in thermodynamics \citep{finitediff}, we use multiple adjacent samples as positive pairs for the reference time step to capture inherent correlations and enhance convergence.

% Instead of relying on data augmentations, which can introduce inductive biases,\gt{this all feels repetitive} CaTT uses adjacent temporal steps as positive pairs, capturing the dependencies between sequences to learn rich feature representations.

To evaluate our learned embeddings, we perform several downstream tasks on embedded data. Our findings demonstrate that CaTT produces representations in the shortest training time, and that downstream tasks performed on data embedded with CaTT outperforms those performed on data embedded with previous state-of-the-art methods.

This paper makes three main contributions: 

\begin{itemize}
    \item Propose CaTT, an unsupervised contrastive representation learning framework for time series that uses N-pair loss to more quickly integrate positive pair selection into the learning process, making the entire process faster and more efficient.

    \item Further adapt the N-pair loss to introduce multiple positive pairs into the normalized temperature-scaled cross-entropy loss (NT-Xent) \citep{simclr}, and adapt it for TS data. Because more pairs are considered for each update, this further accelerates learning. We term this loss \textit{MP-Xent} (multiple positive cross-entropy loss).

    \item Conduct extensive experiments on 16 public datasets and demonstrate superior results compared to state-of-the-art baselines on classification using linear fine-tuning with frozen backbone, semi-supervised classification, forecasting, and anomaly detection.
\end{itemize}

\section{Related Work}
\label{gen_inst}

\textbf{Contrastive representation learning.} Contrastive learning (CL) \citep{hadsell2006} is a widely used self-supervised learning strategy in CV and NLP. Unlike generative models that try to reconstruct inputs, contrastive-based methods aim to learn a data representation by contrasting positive and negative samples, ideally in an unsupervised manner. \citet{npair} introduces the N-pair loss for efficient learning by employing multiple negatives in each batch update. Specifically, \citet{npair} extends triplet loss \citep{triplet} by allowing joint comparison among negative samples. Contrastive predictive coding (CPC) \citep{cpc} learns representation using autoregressive models to predict future time steps in a latent space. A key component of CPC is the introduction of InfoNCE loss, based on noise-contrastive estimation \citep{nce, ncewu} by removing the proximal constraint and using positive pairs. SimCLR \citep{simclr} uses data augmentation and a contrastive loss called NT-Xent that encourages positive pairs (augmented view of the same image) to be closer in the representation space while pushing negative pairs apart. Unlike other contrastive losses that require explicit negative sampling \citep{triplet}, NT-Xent treats all other samples in the batch as negatives, avoiding the need for separate negative sampling strategies, making this loss function more efficient. Our work is built specifically to improve this loss function. \citet{moco} proposes a CL framework that uses a momentum encoder to update the features stored in a dynamic dictionary for stable and consistent feature representation over time. \citet{relic} enforces invariance by adding regularization to the InfoNCE objective. \citet{dcl} further removes the positive pair in the denominator, while in \citet{nnclr}, instead of relying solely on augmentations, uses the nearest neighbor of the current data point in feature space to serve as positive pairs.

\vspace{1em}
\textbf{Contrastive learning in time series.} With the recent traction of CL in CV and NLP, several works in TS representation learning have proposed different methods for sampling positive and negative pairs. \citet{mixup} creates a new augmented sample of a time series and attempts to predict the strength of the mixing components. \citet{tfc} samples positive pairs as time-based and frequency-based representations from the time series signal and introduces a time-frequency consistency framework. \citet{timeclr} introduces dynamic time warping (DTW) data augmentation for creating phase shifts and amplitude changes. To learn discriminative representation across time, TS2Vec \citep{ts2vec} considers the representation at the same time stamp from two views as positive pairs and attempts to enforce feature invariance by jointly optimizing instance-wise CL with temporal CL. InfoTS \citep{infots} focuses on developing criteria for selecting good augmentation in contrastive learning in the TS domain. T-loss \citep{tloss} employs a time-based sample and a triplet loss to learn representation by selecting positive and negative samples based on their temporal distance from the anchor. TNC \citep{tnc} presents temporal neighborhood with a statistical test to determine the neighborhood range that it treats as positive samples. \citet{ncl}, on the other hand, selects neighbors based on both instance-level and temporal-level criteria with a trade-off parameter allowing the model to balance instance-wise distinction with temporal coherence. \citep{clocs} define a positive pair as a representation of transformed instances of the same subject. TS-TCC \citep{tstcc} proposes a method to combine temporal and contextual information in TS using data augmentation to select positives and predict the future of one augmentation using past features of another representation in the temporal contrasting module. CoST \citep{cost} applied CL in learning representation for TS forecasting by having inductive biases in model architecture to learn disentangled seasonal trends. TimeDRL \citep{timedrl} avoids the use of augmentations to eliminate inductive biases and proposes disentangled derivation of timestamp-level and instance-level embeddings from patched time-series data. A more recent work, MF-CLR \citep{mfclr}, presents a method for learning multi-frequency time series representation by adopting a hierarchical mechanism that
traverses different frequencies along the feature dimension.

Unfortunately, these methods involve inefficient sampling in each batch update and fail to leverage the temporal dependencies within the data, ultimately resulting in suboptimal representations. \citet{tnc}, for instance, selects a single positive pair and a single negative pair, akin to \citet{tloss} from subsequences of interest, which can lead to inefficient sampling in every batch update and ignoring the temporal dependencies within the data. \citet{soft} attempts to select more positive pairs from neighboring time steps by creating soft assignments and introducing these to the contrastive loss objectives. However, the effectiveness of this approach is limited by the design of these assignments and the added computational overhead. In our work, we address these drawbacks.

SoftCLT \citep{soft}, in a bid to overcome the issue of ignoring inherent correlations between adjacent timestamps in a sequence, proposes soft assignment to leverage every pair other than the positive pairs by assigning weights to both instance and temporal CL. However, the computation of the soft assignment, which is separate from the contrastive objective, involves distance metrics such as dynamic time warping (DTW) with inherent time and space complexity rendering it unsuitable for real-world time series data with long instances. In comparison to our work, SoftCLT is not an end-to-end CL framework but is built on top of existing contrastive learning approaches to improve performance, adding additional complexity to any inherent shortcomings of the base model.

%\gt{repetitive again} CaTT solves these limitations by incorporating three crucial features. (i) Use adjacent temporal steps as positive pairs and avoiding data augmentations — which can introduce inductive biases; (ii) Adapt N-pair loss for TS to efficiently utilize every sample in a batch as positive and negative pair per update and extend this loss to accept multiple positive pairs to inherent correlations between adjacent instances in a sequence; (iii) Integrate positive and negative pair selection in the contrastive objective in an end-to-end manner to improve speed and efficiency.

\vspace{1em}
\textbf{Feature prediction in representation learning.} Self distillation methods avoid the need for selecting negatives in their training objectives \citep{byol, dino}. They rely on encoding two augmented views and mapping one to the other using a predictor. To avoid mode collapse in self-distillation due to the absence of negatives as in CL, they update one of the encoder weights with the running exponential moving average (EMA) of the other encoder. \citet{simsian} show that the EMA was not necessary in practice, even though it led to a small performance boost. \citet{aleksej} uses the traditional pretext of masked reconstruction to learn feature invariance by a random reconstruction of the masked input of one sensor from another. Masked reconstruction approaches have also produced noteworthy results in forecasting tasks \citep{mm}. TST \citep{tst} attempts to reconstruct masked timestamps using transformers, while PatchTST \citep{patchtst} aims to predict subseries of masked patches to learn local invariant features. SimMTM \citep{simmtm} reconstructs masked time points by aggregating weighted contributions from multiple neighboring points outside the manifold. We compare our method with SimMTM to evaluate the performance between our contrastive learning approach and masked reconstruction.

\section{Proposed architecture: CaTT}
\label{headings}

% Because time-series data is often in the form of waveforms, we assume data is preprocessed using, for instance, Short-Time Fourier Transform (STFT), to produce sub-windows, generating shorter time series {\it instances}. A sequence consists of consecutive, temporally dependent instances, where each instance is closely related to its neighboring ones.

% Our main objective is to learn useful representation from instances of time series data. We assume similarity within nearby instances – that consecutive instances in a sequence have the same class and event labels would not change too often. This condition often holds for time series, which have repeated labels in the temporal dimension.

\begin{figure}[H]
\begin{center}
\includegraphics[scale=0.15]{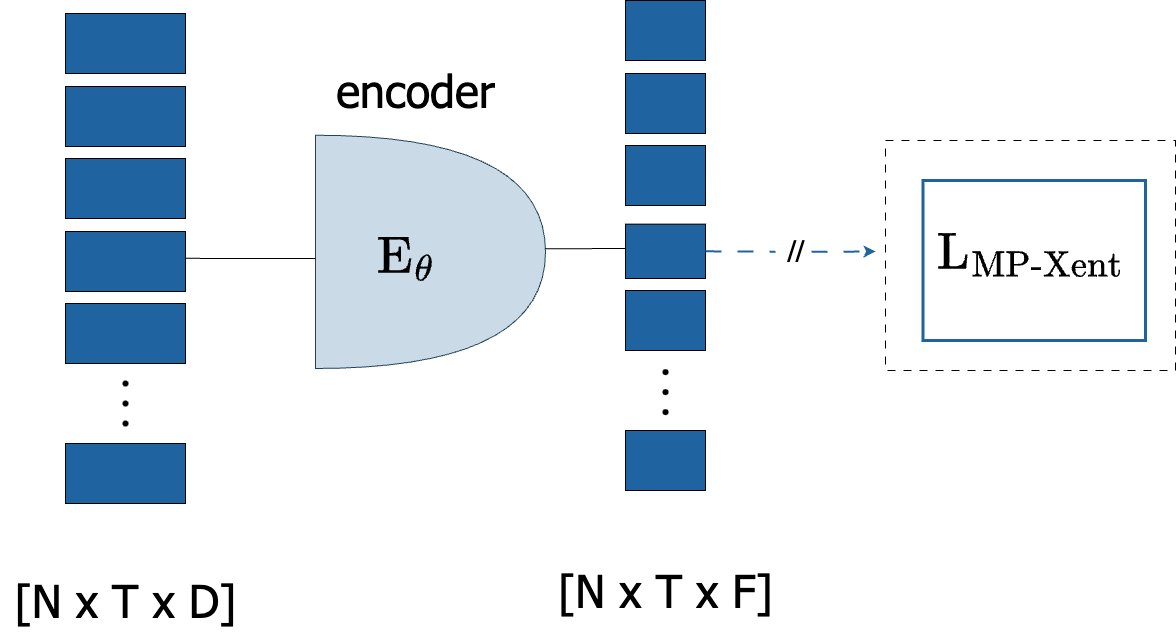}
\end{center}
\caption{Unsupervised representation learning using contrast all the time with MP-Xent objective (CaTT). We train on a batch $N$ of TS instances with sequence length $T$ and feature dimension $D$. The encoder processes this input to generate an embedding vector of $N\times T\times F$ dimension.}
\label{fig:CaTT}
\end{figure}

CaTT learns a mapping function $ f_{\theta}: \bm{x} \rightarrow \bm{z}$, such that given a time series sequence of instances with length $T$, $\bm{x} = \{x_1, x_2,\dots,x_T\}$, where $x_i \in \mathbb{R}^{1 \times D}$, $f_\theta$ projects this series to a representation space $\bm{z} = \{z_1, z_2,\dots,z_T\}$, where $z_i \in \mathbb{R}^{1 \times F}$,  $T$ is the sequence length, $D$ is in the input dimension and $F$ is the dimension of the learned embeddings (Figure \ref{fig:CaTT}). To learn from a training sequence $\bm{x}$ of TS instances, we select an anchor (a single instance), then use adjacent instances as positives and every other sample in the sequence as negatives in the MP-Xent loss, which encourages representations of positive pairs to be similar, and representations of negative pairs to be dissimilar.

%\subsection{Multiple positives cross-entropy (MP-Xent)}
%\label{mploss}

\begin{figure*}[h]
    \begin{center}
        \begin{subfigure}[b]{0.48\textwidth}  % Adjusted width to reduce space
            \centering
            \includegraphics[scale=0.15]{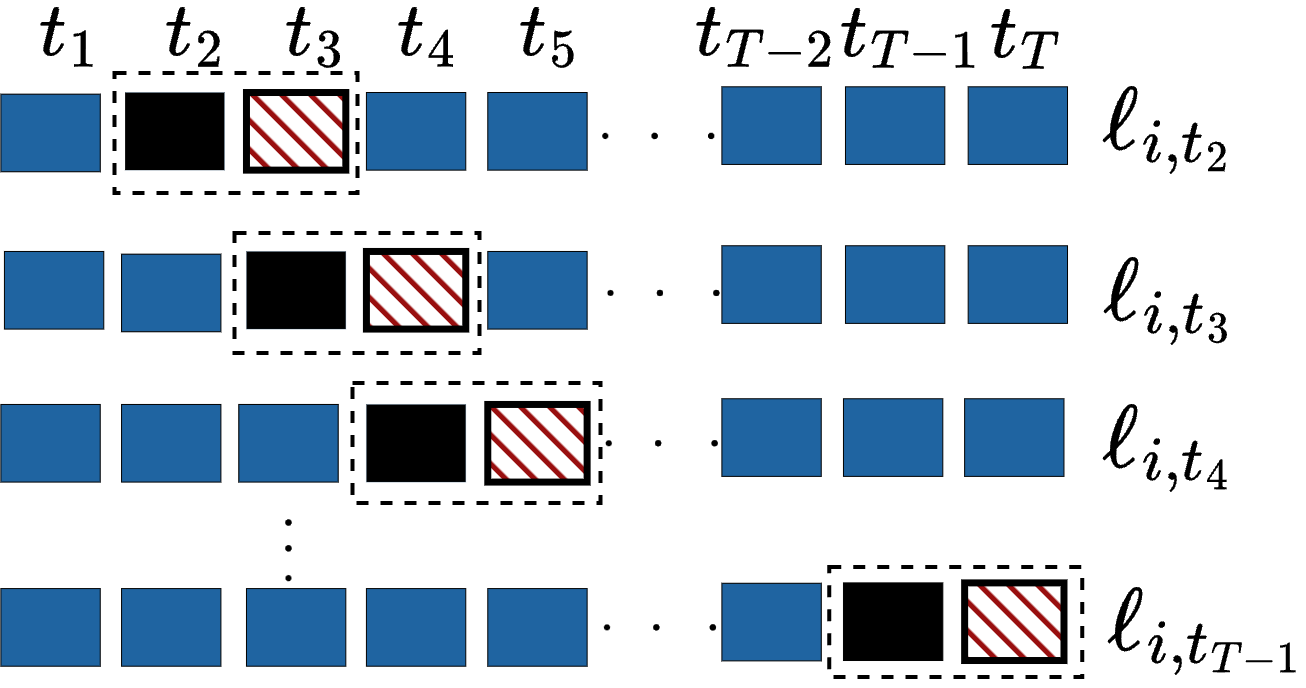}
            \caption{Single positive T-tuple loss}
            \label{fig:image_a}
        \end{subfigure}
        \hspace{0\textwidth}  % Adjust this value to change the space between images
        \begin{subfigure}[b]{0.51\textwidth}  % Adjusted width to reduce space
            \centering
            \includegraphics[scale=0.15]{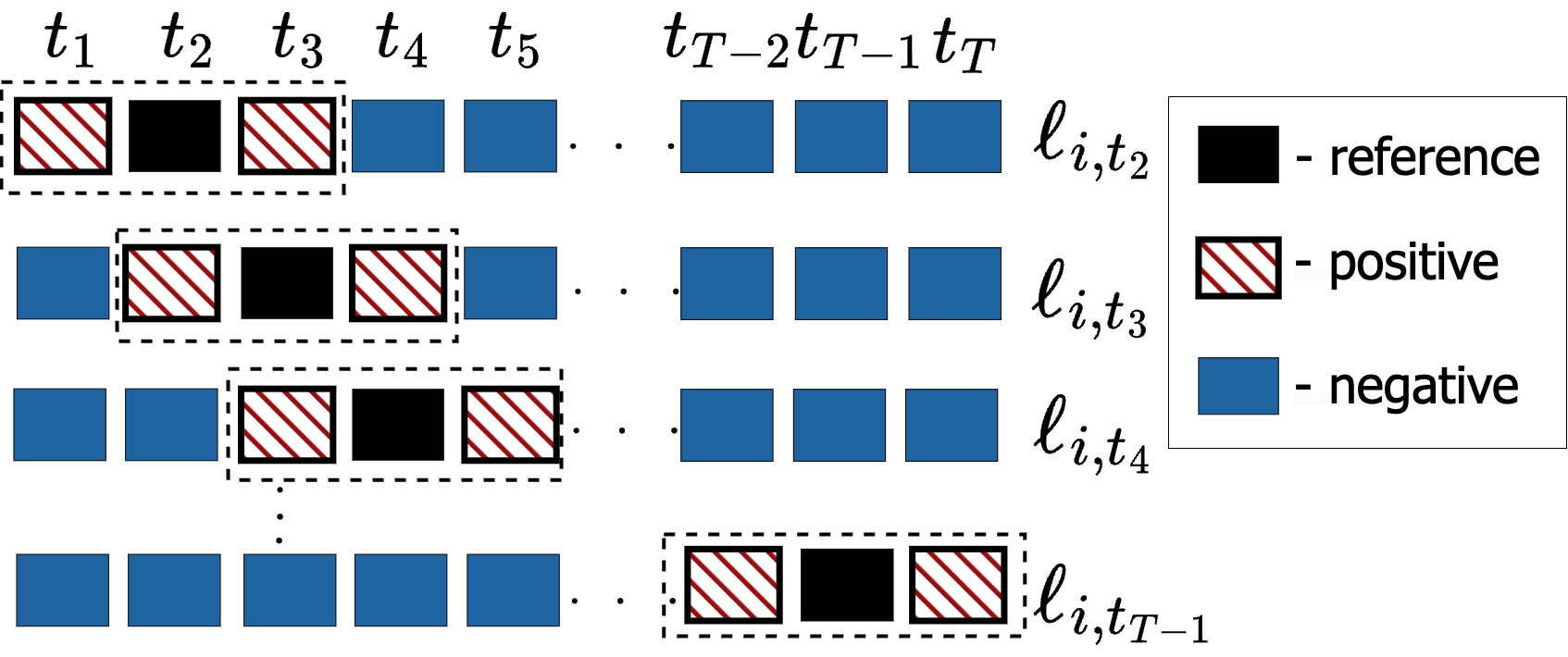} % Replace with your second image file
            \caption{Multiple positives T-tuple loss}
            \label{fig:image_b}
        \end{subfigure}
    \end{center}
    \caption{Positive pairs selection for the contrastive learning objective. The index $i$ refers to the current batch (a) The (N+1) tuple loss \citep{npair} operates on batch N, we adapted this to TS of sequence length T and batch N to obtain NT-tuple losses per batch. (b) We further extend the NT-Xent loss \citep{simclr} by introducing multiple positives for a given reference step from adjacent time steps to enhance convergence. We skip the reference instance $t_1$ and $t_T$ as both losses behave similarly at the edges.}
    \label{fig:combined_images}
\end{figure*}

We now explain our loss function. The MP-Xent loss function is inspired by the N-pair loss as introduced by \citet{npair}, which uses every sample in a batch to compute an (N+1) tuple loss. SimCLR \citep{simclr} builds on this by treating augmented views as positive pairs and all other samples in the batch as negatives. In each batch update, every sample serves as a positive pair at least once. We extend this to TS representation learning by using each instance within a sequence of length $T$. For a batch of size $N$ and sequence $T$, we select each time step as an anchor, adjacent steps as positives, and the rest as negatives, forming an $NT$-tuple loss (Figure \ref{fig:image_a}).

\vspace{2em}
%We train our encoder network $E_\theta(.)$ to learn a representation that clusters similar time series while pushing apart dissimilar time series in space using the MP-Xent objective.
%The encoder $E_\theta(.)$ takes an input $\bm{x}$ such that $\bm{z}$ =  $E_\theta(\bm{x})$ where $\bm{z}$ is the learned feature representation.
Given a single batch $i$, if $z_{i,t}$ and $z_{i,t+1}$ are two consecutive time steps in a sequence of length $T$, with $z_{i,t}$ and $z_{i,t+1}$ $\in \mathbb{R}^{1 \times F}$, equation \ref{eqn:single} shows the NT-Xent loss.

% \begin{equation}
% \ell(i, t) = - \log \frac{\exp(\text{sim}(z_{i,t}, z_{i,t+1}) / \tau)}{\sum_{k=1}^{T} \mathbf{1}_{[k \neq t]} \exp(\text{sim}(z_{i,t}, z_{i,k+1} / \tau)},
% \label{eqn:single}
% \end{equation}

% where $N$ is the batch size, $\tau$ is a temperature parameter \citep{simclr}, and cosine similarity is the similarity score. However, implementing the objective in equation \ref{eqn:single} leads to slow convergence and suboptimal performance. Building upon the principles of the finite difference method \citep{finitediff}, we extend the NT-Xent loss objective in equation \ref{eqn:single} to account for multiple positives for faster convergence and efficient training, as shown in Figure \ref{fig:image_b}. As before, given a reference time step $z_{i,t}$ with adjacent time steps $z_{i,t-1}$ and $z_{i,t+1}$, our MP-Xent loss is as follows.

\begin{equation}
\ell(i, t) = - \log \frac{\exp(\text{sim}(z_{i,t}, z_{i,t+1}) / \tau)}{\sum_{k=1}^{T} \mathbf{1}_{[k \neq t]} \exp(\text{sim}(z_{i,t}, z_{i,k+1} / \tau)},
\label{eqn:single}
\end{equation}

where $T$ is the sequence length, $\tau$ is a temperature parameter \citep{simclr}, and cosine similarity is the similarity score. Implementing the objective in equation \ref{eqn:single} leads to training instability, often resulting in NaN losses and suboptimal performance (Table \ref{tab:ablation}). Building upon the principles of the finite difference method \citep{finitediff}, we extend the NT-Xent loss objective in equation \ref{eqn:single} to account for multiple positives for faster convergence and efficient training, as shown in Figure \ref{fig:image_b}. As before, given a reference time step $z_{i,t}$ with adjacent time steps $z_{i,t-1}$ and $z_{i,t+1}$, our MP-Xent loss is as follows.

\begin{equation}
\ell(i, t) = - \log \frac{P_1(i, t) + P_2(i, t)}{D_1(i, t) + D_2(i, t)},
\label{eqn:main}
\end{equation}
where
\begin{equation}
\begin{aligned}
P_1(i, t) &= \exp(\text{sim}(z_{i,t}, z_{i,t-1}) / \tau), \\
P_2(i, t) &= \exp(\text{sim}(z_{i,t}, z_{i,t+1}) / \tau),
\end{aligned}
\label{eqn:numerator}
\end{equation}

\begin{equation}
\begin{aligned}
D_1(i, t) &= \sum_{k=1}^{T} \mathbf{1}_{[k \neq t, t-1, t+1]} \exp(\text{sim}(z_{i,t}, z_{i,k+1}) / \tau), \\
D_2(i, t) &= \sum_{l=1}^{T} \mathbf{1}_{[l \neq t, t-1]} \exp(\text{sim}(z_{i,t-1}, z_{i,l}) / \tau).
\end{aligned}
\label{eqn:denominator}
\end{equation}

For the entire sequence length $T$ and batch $N$, we have an \textit{NT} tuple loss per update, making our training very efficient.

\begin{equation}
L_{MP-Xent} = \frac{1}{NT} {\sum_{i=1}^{N} {\sum_{t=1}^{T} \ell(i, t)}}
\label{eqn:mp-xent}
\end{equation}

Extending the N-pair loss~\cite{npair} to account for multiple positives as in equation \ref{eqn:mp-xent} while maintaining efficient training through matrix operations is non-trivial. A naive approach would involve selecting multiple positives by slicing the temporally adjacent steps for each time step in sequence; however, this incurs high computational complexity.

To address this, our method efficiently integrates positive selection and MP-Xent loss computation by adapting the N-pair loss formulation to accommodate two positives, while preserving matrix-based efficiency. This adaptation enables each sample in a batch and time step to contribute to an $NT$-tuple loss, as detailed in Appendix~A.

Algorithm (1) is derived through mathematical induction by observing the pattern in the similarity matrix when considering multiple positives.

\begin{algorithm}[H]
\caption{MP-Xent Loss with Multiple Positives for Batch of Sequences}
\begin{algorithmic}[1]
\REQUIRE Batch of time series $\bm{X} \in \mathbb{R}^{N \times T \times D}$
\STATE Encode each time step: $\bm{Z} = f_\theta(\bm{X}) \in \mathbb{R}^{N \times T \times F}$
\STATE Reshape $\bm{Z}$ to 2D matrix: $\bm{Z}_{\text{flat}} \in \mathbb{R}^{(N \cdot T) \times F}$
\STATE Compute similarity matrix: $\bm{S} = \bm{Z}_{\text{flat}} \cdot \bm{Z}_{\text{flat}}^\top \in \mathbb{R}^{(NT) \times (NT)}$
\STATE Extract the lower diagonal elements of $S$
\STATE The positive pairs are the sum of the shifted left and right of the lower diagonal elements (numerator)
\STATE The negatives is the sum of all elements in the similarity matrix (except the last two columns along each row).
\STATE Subtract a combination of two lower diagonal slices from the sum to produce the denominator (Equation ~\ref{eqn:main}.
\STATE Compute the $NT$-tuple loss using the numerator and denominator (Equation~\ref{eqn:mp-xent})
\end{algorithmic}
\end{algorithm}

% \begin{figure}[h!]
% \begin{center}
% \includegraphics[scale=0.15]{CaTT.png}
% \end{center}
% \caption{Unsupervised representation learning using contrast all the time with margin (CaTT-M). We train on TS instances of length T and feature dimension D. (Right to left): We mask random features from the time series instances and use this as input to the encoder. The encoder processes this mask input to generate an embedding vector.}

\section{Experiments}
\label{sec4}

To evaluate the performance of CaTT we use three benchmark datasets to assess the quality of the learned embeddings. We compare our approach against state-of-the-art baselines for time series representation learning on classification using linear fine-tuning with a frozen backbone, semi-supervised classification, and forecasting. We demonstrate that CaTT outperforms other approaches in building semantically meaningful representations, in less time. Finally, we perform an ablation study to highlight the effects of the different components of our CaTT models.

CaTT assumes that consecutive instances in a sequence are similar and have the same class. This typically holds for real-world datasets but generally does not for datasets in the UCR \citep{ucr} or UEA \citep{uea} archives with short instances. Therefore, we follow conventions in works like \citet{tfc, timedrl, tnc, tstcc, simmtm, aleksej} by using long instances of real-world datasets. Due to the nature of medical time series data, we preprocess the data by applying the Short-Time Fourier Transform (STFT) to sub-windows, generating shorter time series instances. For the remainder of this section, we refer to an instance as one of these preprocessed blocks of data. A sequence T consists of consecutive, temporally dependent instances, where each instance is closely related to its neighboring ones. This preprocessing step enables the model to better capture the temporal structure and patterns inherent in the medical signals. We evaluate our model on classification using three public real world datasets on human activity recognition, electrical activity of the heart, and sleep stage classification. Table \ref{tab:dataset} shows the summary statistics of these datasets.

\begin{table}[ht]

\centering

\caption{Summary of dataset distributions used in the classification experiments after applying STFT. An instance is a preprocessed block of TS. For the \textsc{Harth} each instance is 1 second while for the \textsc{SleepEeg} and \textsc{Ecg} an instance is 2 and 6 seconds respectively.}

\label{sample-table}
\vskip 0.15in

\resizebox{0.48\textwidth}{!}{ 
  \begin{tabular}{lcccccc}
    \toprule
    & \# Instance & Sequence length & Dimension & Classes \\
    \midrule

    \textbf{\textsc{Harth}} & 1,270,087 & 119 & 156 & 12  \\
    \textbf{\textsc{SleepEeg}} & 478,600  & 200 & 90 & 5  \\
    \textbf{\textsc{Ecg}} & 255,200 & 200 & 1502 & 4  \\
    
    \bottomrule
  \end{tabular}
}
\vskip -0.1in
\label{tab:dataset}
\end{table}

\textsc{\textbf{Harth} -} This is a human activity recognition (HAR) dataset \citep{harth} that contains recordings from 22 participants, each wearing two 3-axial Axivity AX3 accelerometers for approximately 2 hours in a free-living setting at a sampling rate of 50Hz. This dataset comprises 12 distinct classes of varying human activities \textit{(standing, lying, walking, shuffling, running, sitting, stairs - ascending and descending, and four different cycling positions)}. We preprocess the signal by applying a short-time Fourier transform (STFT) using a one-second Hann window \citep{hann, aleksej} with a half-second overlap. We then concatenate the activities from all 22 subjects to build a continuous time series, resulting in a spectrogram with 1,270,087 instances and 156 feature dimensions. During our unsupervised representation learning, for each iteration, we use a sequence length of 119 instances, corresponding to 60 seconds, and encode the representations in a 320-dimensional space.

\textsc{\textbf{SleepEeg} -} This dataset \citep{sleepeeg} contains 153 whole-night electroencephalography (EEG) sleep recordings from 82 healthy subjects, sampled at 100 Hz. We preprocessed the dataset from \citet{tfc}, we apply a short-time Fourier transform (STFT) on the raw signal to produce 478,600 instances with a feature dimension of 90. Each sample corresponds to one of the five sleep stages: Wake (W), Non-Rapid Eye Movement (N1, N2, N3), and Rapid Eye Movement (REM). In our training, we use a sequence length of 200 and output representations in a 320-dimensional space.

\textsc{\textbf{Ecg} -} We use the MIT-BIH Atrial Fibrillation dataset \citep{ecg}, which includes 25 long-term electrocardiogram (ECG) recordings of human subjects with atrial fibrillation, each with a duration of 10 hours. The dataset contains two ECG signals, each sampled at 250 Hz, with annotations marking the different rhythms: atrial fibrillation (A), atrial flutter (F), AV junctional rhythm (AV), and all other rhythms. Similar to the \textsc{Harth} dataset, we apply a short-time Fourier transform (STFT) with a six seconds Hann window and a 3-seconds overlap, producing a total of 255,200 instances with a feature dimension of 1502. Finally, we select 200 instances, as sequence length. The learned representations are encoded in a 320-dimensional space. This dataset is particularly useful for evaluating how our proposed method performs on imbalanced data, as the atrial (A) rhythm and ``all other rhythms" account for more than 99\% of the entire dataset.

\vspace{1em}
\textbf{Baselines.} We compare our model with eight state-of-the-art approaches in time series representation learning: InfoTS \citep{infots}, SimMTM \citep{simmtm}, TNC \citep{tnc}, TS2Vec \citep{ts2vec}, Soft \citep{soft}, TimeDRL \citep{timedrl}, CoST \citep{cost} and MF-CLR \citep{mfclr}. InfoTS maximizes agreement between representations of the same subseries through temporal augmentations. SimMTM constructs masked time points by aggregating weighted contributions from multiple neighboring points. TNC learns by contrasting data points within the same neighborhood against those from different neighborhoods. TS2Vec captures both global and temporal dependencies by contrasting time series across different scales and timesteps, Soft improves on existing CL approaches by precomputing soft assignments. TimeDRL eliminates the need for augmentations to remove inductive biases and introduces a disentangled approach for deriving timestamp-level and instance-level embeddings from patched time-series data. CoST employs a two-step approach to TS forecasting by learning disentangled seasonal trends, while MF-CLR presents a method for learning multi-frequency time series representation by adopting a hierarchical mechanism. To ensure a fair comparison, all models were trained using the same preprocessing pipeline and hyperparameters. Specifically, we employed the AdamW optimizer with a learning rate of \(1 e^{-3}\) and a batch size of 8. Additionally, to eliminate any performance differences arising from variations in model architecture, we use the same backbone encoder network across all baselines. We aim to compare the learning frameworks independent of the choice of encoder. To this end, we selected a simple CNN architecture to assess how effectively each framework can leverage the limited capacity of a basic encoder to learn meaningful representations. Consequently, we substituted the dilated CNN unique to the TS2Vec encoder with a regular  1D CNN (refer to the appendix for more details on each baseline and implementation).

% \vspace{1em}
\textbf{Network architecture}. We use a simple convolutional neural network (CNN) architecture as our feature extractor backbone (see appendix) in our encoder \( E_\theta(\cdot) \). The CNN network consists of 3 blocks of 1D convolution with a kernel size of 1, followed by batch normalization and ReLU activation, with an embedding dimension of 32. Since our focus is on developing a loss function, we use the same architecture for all baselines in Section \ref{sec4}. To preprocess the time series signal for our encoder, we perform a short-time Fourier transform (STFT) on the signal to obtain input with dimension \( B \times T \times D \), where \( B \) is the batch size, \( T \) is the sequence length, and \( D \) is the input dimension.

% \vspace{1em}
\textbf{Pretraining details}. We perform a 50-50 train test split on the \textsc{Harth} and an 80-20 train test split on both the \textsc{SleepEeg} and \textsc{Ecg} datasets. For the pretraining of all models, we maintain the same hyperparameters. Specifically, for the forecasting task, the batch size is set to 8, and the learning rate is 0.001. The number of optimization iterations is 200 for datasets smaller than 100,000 and 600 otherwise. The representation dimension is fixed at 320, following \citet{ts2vec}. We use the same setup for the classification tasks, except for setting the number of training iterations on the \textsc{Ecg} with a size of approximately 150,000 to 200 iterations. We train all models on an NVIDIA V100 GPU.

\subsection{Linear evaluation with frozen backbone}

Our main goal is to learn representations that are useful in downstream tasks. With that in mind, we train a linear classifier on top of the learned representations to assess how well the learned features generalize to the task of interest when used by a simple classifier. The results in Table \ref{tab:linear} demonstrate that CaTT consistently outperforms other baselines across all evaluation metrics - accuracy, F1 score, precision, and recall — on all three datasets. Furthermore, CaTT achieves this superior performance with lower computational time.

\begin{table}[ht]
\centering
\caption{ Comparison with state-of-the-art methods on linear evaluation with a frozen backbone. We compare CaTT with baselines and a randomly initialized encoder (\textit{Random Init.}) on frozen evaluation. We train a linear classifier on top of the features from the encoder using the entire train set and evaluate the test set. CaTT achieved the best performance on all three datasets despite having the shortest training time.}
\vskip 0.15in
\resizebox{0.47\textwidth}{!}{ 

  \tiny % Reduce the font size
  \begin{tabular}{llccccc}
    \toprule
    \textbf{Datasets} & \textbf{Models} & \textbf{Accuracy} &  \textbf{F1 score} & \textbf{Precision} & \textbf{Recall} & \textbf{Time (s) $\downarrow$}\\
    \midrule

    \multirow{8}{*}{\textbf{\textsc{Harth}}} 
    & Random Init.  &86.69±0.50  &0.84±0.01  &0.55±0.01 &0.45±0.02&-\\

    & TNC        &89.03±0.03  &0.88±0.00  &0.70±0.02  &0.55±0.01 &13.86\\
    
    & InfoTS         &90.12±0.13  &\textbf{0.89±0.00}  &\textbf{0.71±0.02}  &0.56±0.01 &186.50\\
    & CoST         &88.68±0.41  &0.88±0.00  &0.64±0.01  &0.56±0.00 &12.61\\
    & SimMTM    &68.14±4.83  &0.60±0.08  &0.34±0.04  &0.24±0.05 &245.67\\
    & TimeDRL     &71.92±2.37  &0.63±0.04  &0.37±0.04  &0.28±0.01 &20.60\\
    & TS2Vec     &87.59±0.23  &0.85±0.00  &0.58±0.01  &0.49±0.00 &34.96\\

    & TS2Vec + \textit{SoftCLT}     &88.07±0.13  &0.86±0.00  &0.62±0.03  &0.50±0.01 &37.65\\
    
    & MF-CLR        &77.07±5.38  &0.73±0.07& 0.52±0.04  &0.37±0.05 &2322.75\\

    & CaTT        &\textbf{90.35±0.05}  &\textbf{0.89±0.00}&\textbf{0.71±0.02}  &\textbf{0.59±0.00} &\textbf{6.12}\\

    \midrule
    
    \multirow{8}{*}{\textbf{\textsc{SleepEeg}}} 
    & Random Init.  &46.89±11.34  &0.44±0.08  &0.36±0.03  &0.32±0.04&- \\

    & TNC        &55.13±0.50  &0.54±0.01  &0.33±0.01  &0.36±0.02 &10.14\\
    
    & InfoTS         &43.32±9.66  &0.42±0.07  &0.41±0.05  &0.35±0.03 &260.35\\
    & CoST         &56.63±2.56  &0.55±0.00  &0.38±0.02  &0.35±0.04 &11.89\\
    & SimMTM    &57.08±0.68  &0.50±0.01  &0.31±0.03  &0.25±0.01 &1739.66\\
    & TimeDRL     &57.30±2.22  &0.48±0.02  &0.36±0.09  &0.24±0.02 &13.20\\
    & TS2Vec     &51.18±1.12  &0.50±0.00  &0.29±0.01  &0.35±0.02 &34.42\\

    &  TS2Vec + \textit{SoftCLT}     &53.60±2.44  &0.52±0.01  &0.34±0.03  &0.37±0.02 &36.48\\

    & MF-CLR        &57.07±1.89  &0.48±0.02&0.27±0.06  &0.27±0.01 &2318.26\\
    
    & CaTT        &\textbf{62.39±1.21}  &\textbf{0.60±0.01}&\textbf{0.47±0.00}  &\textbf{0.41±0.01} &\textbf{6.36}\\
    
    \midrule

    \multirow{8}{*}{\textbf{\textsc{Ecg}}} 
   & Random Init.  &65.48±1.44  &0.64±0.01  &0.35±0.01  &0.34±0.01 &- \\
    & TNC        &75.86±2.00  &0.75±0.02  &0.39±0.00  &0.38±0.01 &4.86\\
    & InfoTS         &61.94±3.19  &0.59±0.05  &0.34±0.01  &0.32±0.01 &76.42\\
    
    & CoST         &69.37±9.55  &0.67±0.13  &0.38±0.02  &0.36±0.04 &11.27\\
    & SimMTM    &71.89±2.27  &0.71±0.03  &0.37±0.00  &0.36±0.01 &594.86\\
    & TimeDRL     &59.51±6.93  &0.55±0.10  &0.33±0.02  &0.31±0.03 &13.21\\
    & TS2Vec     &77.72±1.27  &0.77±0.01  &0.40±0.00  &0.39±0.01 &11.42\\

    & TS2Vec + \textit{SoftCLT}      &76.37±3.88  &0.76±0.04  &0.39±0.02  &0.39±0.02 &16.70\\

    & MF-CLR        & 52.69±2.29  &0.45±0.06&0.31±0.01  &0.28±0.01 &3056.11\\
    
    & CaTT        &\textbf{80.18±3.80}  &\textbf{0.81±0.03}&\textbf{0.41±0.01}  &\textbf{0.41±0.02} &\textbf{4.47}\\
    
    \bottomrule
  \end{tabular}
    }
% \vskip -0.1in
\label{tab:linear}
\end{table}

Table \ref{tab:linear} shows the \textsc{Harth} dataset where CaTT achieves the highest accuracy (90.35\%) and F1 score (0.89) with the lowest computational time (6.12s), outperforming computationally intensive methods such as MF-CLR (2322.75), SimMTM (245.67 seconds) and InfoTS (186.50 seconds). The \textsc{SleepEeg} dataset presents a more challenging task, as indicated by lower overall scores across models. Nonetheless, CaTT leads with 62.39\% accuracy, an F1 score of 0.60, and efficient runtime (6.36s), significantly faster than MF-CLR (2318.26), SimMTM (1739.66s) and InfoTS (260.35s). On the \textsc{Ecg} dataset, CaTT achieves an accuracy of 80.18\% and an F1 score of 0.81, outperforming the second-best model, TS2Vec, which scores 77.72\% in accuracy and 0.77 in F1. Interestingly, while while TS2Vec + \textit{SoftCLT} shows marginal improvements in some metrics over TS2Vec, it still falls short of CaTT’s performance. Additionally, CaTT achieves its results with the lowest computational time of 4.47 seconds, demonstrating its efficiency. These results establish CaTT as a highly effective and practical framework for activity recognition and physiological signal analysis.

\subsection{Semi-Supervised classification}

To augment our experiments on downstream tasks and provide insight into method behaviors, we perform semi-supervised classification on sparsely labeled datasets to evaluate the quality of the learned embeddings. All models are pre-trained, followed by fine-tuning a linear classifier on subsets of the training data (5\%, 10\%, and 20\%) to predict labels on the evaluation set. The average accuracy and standard deviation over 15 runs are presented in Table \ref{tab:knn}.

Our model, CaTT, demonstrates consistently strong performance in the \textsc{Harth}, \textsc{SleepEeg}, and \textsc{Ecg} datasets. In particular, CaTT achieves the highest accuracy in most settings, particularly in 20\% labeled data, outperforming competing methods such as TS2Vec, CoST, and InfoTS. For example, on the \textsc{Harth} dataset, CaTT achieves 88.93\% accuracy with a minimal deviation of ±0.73, highlighting its stability. On the \textsc{SleepEeg}, CaTT achieves superior performance at 5\%, 10\%, and 20\% labeled data, reaching 59.99\%, 61.82\%, and 61.95\% accuracy, respectively, surpassing other approaches like CoST and InfoTS. Similarly, on the \textsc{Ecg} dataset, CaTT demonstrates particularly strong results across all label proportions, including the best performance at 20\% labeled data with 77.43\% accuracy, highlighting its robustness in capturing meaningful representations even with limited labeled data.

\begin{table*}[h]
\centering
\caption{Semi-supervised classification performance over 15 runs, showing mean accuracy (± standard deviation) across datasets (\textsc{Harth}, \textsc{SleepEeg}, and \textsc{Ecg}) with varying label fractions (5\%, 10\%, 20\%). Results compare CaTT against baseline methods, highlighting the effectiveness of CaTT in all settings.}

\resizebox{1\textwidth}{!}{ 
  \begin{tabular}{lcccccccccc}
    \toprule
    & \multicolumn{3}{c}{\textbf{\textsc{Harth}}} & \multicolumn{3}{c}{\textbf{\textsc{SleepEeg}}} & \multicolumn{3}{c}{\textbf{\textsc{Ecg}}} \\
    \cmidrule(lr){2-4} \cmidrule(lr){5-7} \cmidrule(lr){8-10} % Half-rule lines for precision and recall columns
    & 5\% & 10\% & 20\% & 5\% & 10\% & 20\% & 5\% & 10\% & 20\% \\
    
    \midrule

    TNC  & 84.96±2.05 & 86.78±0.89 & 87.94±0.23 & 54.71±2.57 & 54.02±2.23 & 52.70±3.10 & 63.87±4.18 & 71.65±6.03 & 73.32±7.33\\
    InfoTS  & \textbf{87.39±0.92} & \textbf{88.16±0.64} & 88.69±0.64 & 40.34±11.25 & 37.80±11.51 & 43.50±9.26 & 52.30±8.30 & 55.55±6.97 & 57.16±6.39\\
    
    CoST  & 86.84±0.70 & 87.59±0.43 & 88.24±0.28 & 55.13±2.83 & 53.96±3.52 & 55.70±2.99 & 62.68±3.51 & 65.90±2.78 & 66.90±4.79\\
    SimMTM  & 59.35±2.13 & 60.07±1.37 & 61.06±1.37 & 57.39±1.58 & 57.72±1.61 & 56.93±2.04 & 56.63±5.93 & 60.90±5.62 & 65.86±4.54\\
    TimeDRL  & 56.95±1.84 & 62.00±2.52 & 66.70±1.39 & 55.60±4.54 & 56.30±3.52 & 56.02±3.56 & 49.85±2.45 & 50.50±2.18 & 49.84±1.46\\
    TS2Vec  & 79.18±2.27 & 83.98±0.70 & 85.36±0.51 & 57.80±1.39 & 55.05±2.84 & 51.85±2.11 & 58.46±8.32 & 71.11±6.96 & 74.23±6.88\\

    TS2Vec + \textit{SoftCLT} & 83.08±1.47 & 85.02±1.02 & 86.36±0.63 & 56.35±1.02 & 54.24±2.09 & 54.65±2.34 & 63.24±9.57 & 69.30±8.73 & 75.32±7.36 \\

    % \midrule
    MF-CLR  & 62.06±1.54 & 64.47±1.48 & 66.50±0.52 & 43.81±7.26 & 46.27±10.51 & 48.26±8.40 & 59.08±4.56 & 57.84±6.19  & 62.82±3.35\\
    
    CaTT  & 84.74±2.26 & 86.90±1.58 & \textbf{88.93±0.73} & \textbf{59.99±2.33} & \textbf{61.82±1.48} & \textbf{61.95±1.70} & \textbf{73.37±10.10} & \textbf{73.84±6.58 } & \textbf{77.43±4.21}\\

    \bottomrule
  \end{tabular}
}
\label{tab:knn}
\end{table*}

\subsection{Visualization of learned representations}

\begin{figure}[ht!]
\centering
\includegraphics[scale=0.2]{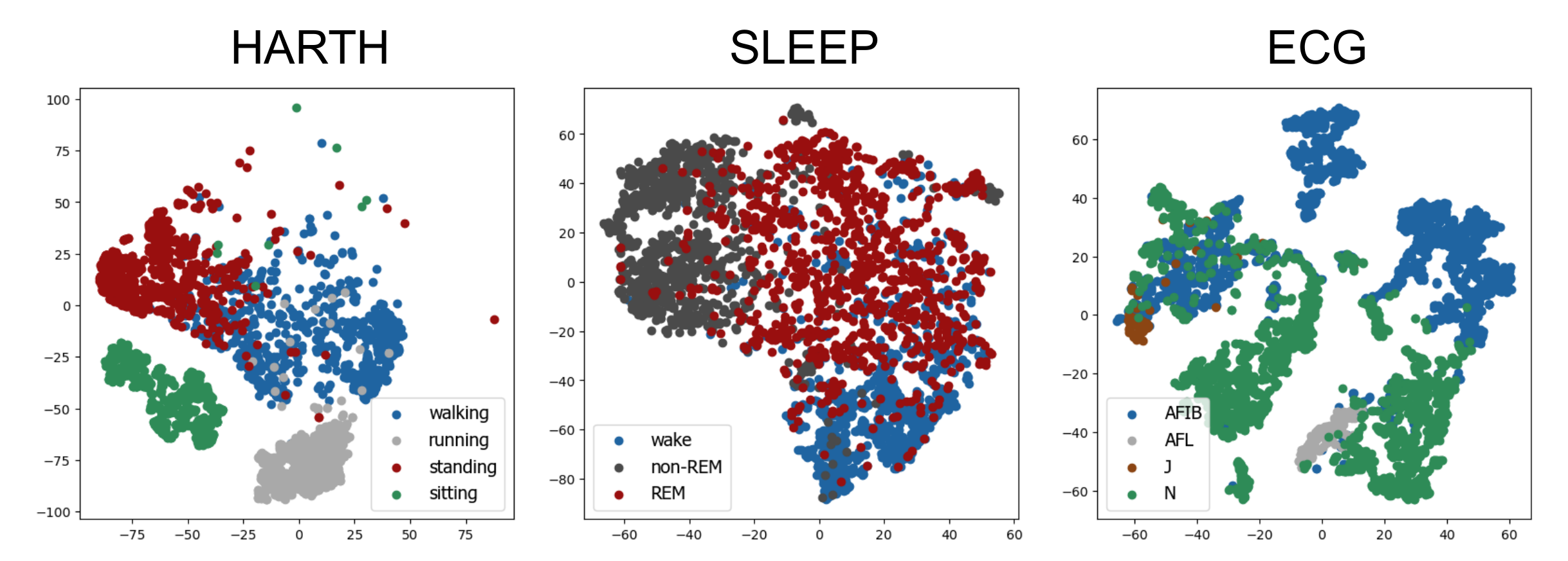}
\caption{t-SNE visualization of the learned embeddings by CaTT on random instances on the \textsc{Harth}, \textsc{SleepEeg} and \textsc{Ecg}. For the \textsc{SleepEeg} each instance (data point) spans 2 seconds, while for the \textsc{Harth} and \textsc{Ecg} each instance is 1 and 6 seconds respectively. The same colour represents an instance from the same class.}
\label{fig:sleepfeaures}
\end{figure}

In addition to our representation being useful in downstream tasks, we also want to learn compact and semantically meaningful representations. We seek to understand how consistently the learned representation clusters similar instances together, despite not having access to this information during training. This is a good indicator of whether the representations are meaningful. To that end, we visualize a random subset from that test set. Figure \ref{fig:sleepfeaures} shows a t-SNE plot of the learned representation from CaTT on all three datasets. Our CaTT model embeds instances into well-defined, semantically meaningful clusters.

\subsection{Time series forecasting}

To evaluate the timestamp-level embeddings of CaTT models, we conducted experiments on short-term and long-term forecasting across four real-world public benchmark datasets: ETT (Electricity Transformer Temperature) \citep{ETT} consists of two hourly-level datasets (ETTh) and one 15-minute-level dataset (ETTm), measuring six power load features and the target, oil temperature, and Weather\footnote{https://www.ncei.noaa.gov/data/local-climatological-data/} which is an hourly-level dataset containing 11 climate features.

\begin{figure}[H]
\hspace*{-0.45cm}  % Adjust the value to move the figure left
\includegraphics[scale=0.65]{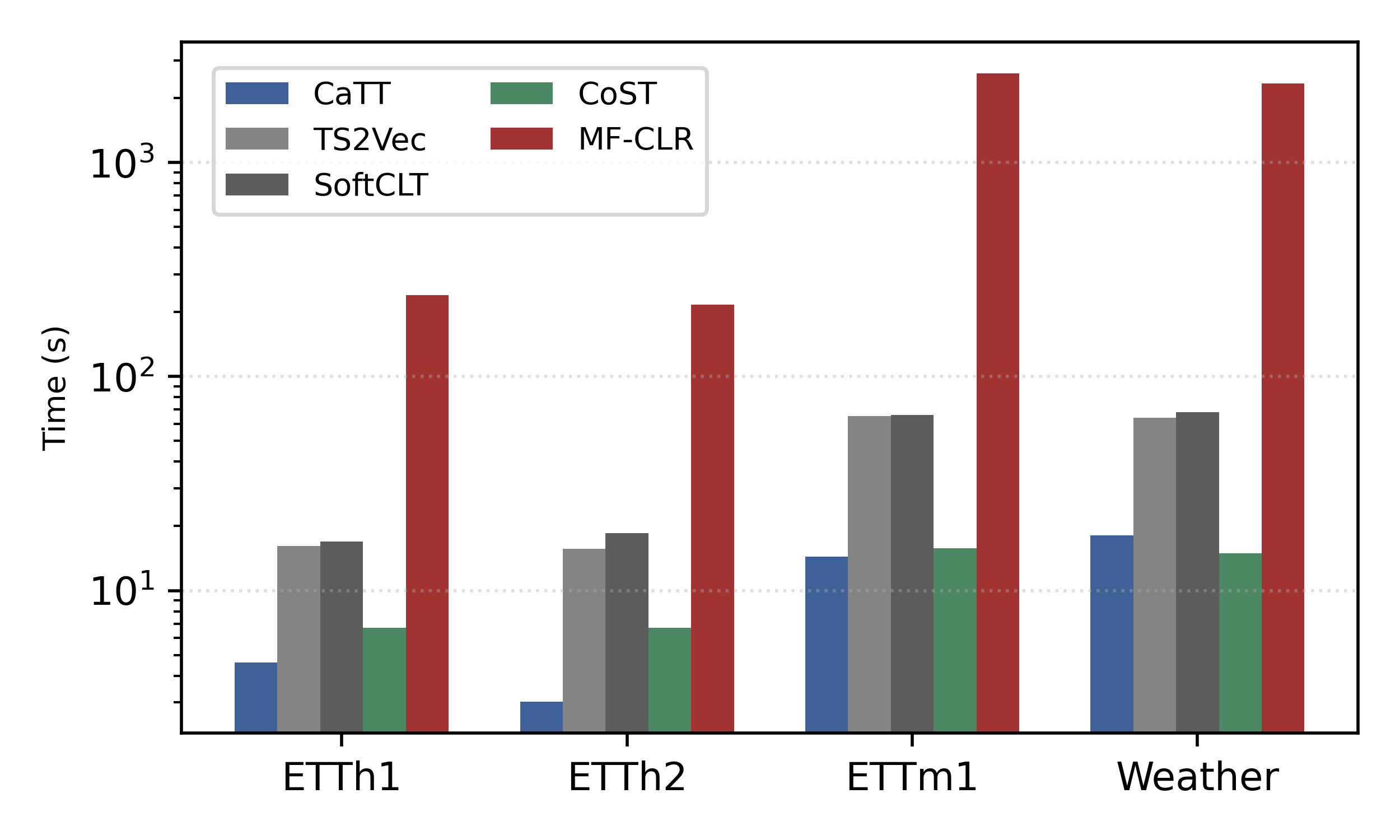}
\caption{Unsupervised pretraining time (in seconds) for short-term and long-term forecasting.}
\label{fig:pretime}
\end{figure}

The results are summarized in Table \ref{forecast-univar}. Reporting Mean Squared Error (MSE) and Mean Absolute Error (MAE) for prediction horizons (H) from 24 to 720, we compared CaTT against TS2Vec, Soft, CoST, and MF-CLR baselines known for good forecasting performance. For a fair comparison, we use the same dilated convolutional backbone feature extractor introduced by TS2Vec across all baselines. CaTT consistently delivered competitive results, matching MF-CLR and outperforming CoST and TS2Vec on ETTh$_1$. Similarly, CaTT performs equally well with CoST while outperforming TS2Vec, Soft, and MF-CLR on the ETTm$_1$ long Horizon forecasting. Moreover, CaTT demonstrates significant advantages on the Weather dataset, where it achieves the lowest average MAE, outperforming TS2Vec, CoST, Soft, and MF-CLR. Across all datasets, the average MSE and MAE highlight the overall efficiency of CaTT, with averages of 0.088 and 0.211, respectively, surpassing CoST (MSE = 0.092, MAE = 0.217), SoftCLT (MSE = 0.099, MAE = 0.229) and MF-CLR (MSE = 0.096, MAE = 0.224). To complement the forecasting performance, we analyze the unsupervised pretraining time in Figure \ref{fig:pretime}. CaTT (40.12) demonstrates remarkable efficiency, requiring significantly less time than MF-CLR (5392.83), TS2Vec (161.07), Soft (169.66) and slightly lesser than CoST (44.08). These results underscore the robustness and efficiency of CaTT. CaTT not only delivers competitive forecasting accuracy but also offers substantial computational advantages during pretraining. This balance of performance and efficiency makes CaTT particularly well-suited for real-world time series forecasting tasks, where scalability and speed are critical.

% \begin{table}[H]
% \centering
% \caption{Unsupervised pretraining time (in seconds) for short-term and long-term forecasting.}
% \vskip 0.15in

% \resizebox{0.45\textwidth}{!}{ 
%   \begin{tabular}{lcccccc}
%     \toprule
%     & \multicolumn{5}{c}{\textsc{Time (s)}} \\
%     \cmidrule(lr){2-6} 
    
%     & CaTT & TS2Vec & TS2Vec + \textit{SoftCLT} & CoST & MF-CLR \\
%     \midrule

%     ETTh1 & \textbf{4.62} & 16.12 & 16.93 & 6.70 & 239.33\\
%     ETTh2 & \textbf{3.02} & 15.63 & 18.50 & 6.69 & 216.32\\
%     ETTm1 & \textbf{14.43} & 65.37 & 66.11 & 15.80 & 2604.35\\
%     Weather & 18.05 & 63.95 & 68.12 & \textbf{14.89} & 2332.83\\
%    \midrule
%     Total & \textbf{40.12} & 161.07 & 169.66 & 44.08 & 5392.83\\

%     \bottomrule
%   \end{tabular}
% }
% \label{tab:fore_time}
% \vskip -0.1in
% \end{table}

\begin{table}[H]
  \centering
  \caption{Short-term and long-term forecasting results across multiple datasets (ETTh$_1$, ETTh$_2$, ETTm$_1$, Weather) and forecast horizons (H = 24, 48, 168, 336, 720). Performance is evaluated using Mean Squared Error (MSE) and Mean Absolute Error (MAE). CaTT demonstrates competitive accuracy compared to baseline methods, with the best results highlighted in bold.}
  \vskip 0.15in
  \scalebox{0.63}{
  \begin{tabular}{lccccccccccccc}
  \toprule
         &  & \multicolumn{2}{c}{CaTT} &
         \multicolumn{2}{c}{TS2Vec} &
         \multicolumn{2}{c}{TS2Vec + \textit{SoftCLT}} &
         \multicolumn{2}{c}{CoST} &
         \multicolumn{2}{c}{MF-CLR}\\
    \cmidrule(r){3-4} \cmidrule(r){5-6} \cmidrule(r){7-8} \cmidrule(r){9-10} \cmidrule(r){11-12} \cmidrule(r){13-14} 
    Dataset & H & MSE & MAE & MSE & MAE & MSE & MAE & MSE & MAE & MSE & MAE\\
    \midrule
    \multirow{5}*{ETTh$_1$}
    & 24 & 0.040 & 0.152 & 0.040 & 0.153 & 0.040 & 0.153 & 0.041 & 0.152 & \textbf{0.039} & \textbf{0.149}  \\
    
    & 48 & 0.063 & 0.193 & 0.063 & 0.193 & 0.063 & 0.192& 0.066 & 0.196 & \textbf{0.062} & \textbf{0.189} \\
    
    & 168 & \textbf{0.122} & \textbf{0.269} & 0.145 & 0.296 & 0.144 & 0.296 & 0.126 & 0.273 & 0.142 & 0.297 \\
    
    & 336 & \textbf{0.143} & \textbf{0.296} & 0.164 & 0.321 & 0.164 & 0.321& 0.150 & 0.305 & 0.165 & 0.323 \\
    
    & 720 & 0.189 & 0.354 & \textbf{0.174} & \textbf{0.340} & \textbf{0.174} & 0.342 & 0.188 & 0.354 & 0.201 & 0.371 \\
    
    \midrule
    
    \multirow{5}*{ETTh$_2$}
    & 24 & 0.095 & 0.235 & 0.090 & 0.227 & \textbf{0.089} & \textbf{0.226}& 0.091 & 0.231 & 0.093& 0.230  \\
    & 48 & 0.130 & 0.279 & 0.126 & 0.275 & \textbf{0.125} & \textbf{0.274}& 0.127 & 0.277 & 0.133 & 0.279 \\
    & 168 & \textbf{0.190} & \textbf{0.348} & 0.200 & 0.357 & 0.201 & 0.357& 0.206 & 0.359 & 0.200 & 0.352 \\
    & 336 & \textbf{0.196} & \textbf{0.357} & 0.205 & 0.365 & 0.206 & 0.366 & 0.206 & 0.363 & 0.206 & 0.363 \\
    & 720 & \textbf{0.201} & 0.365 & 0.205 & 0.370 & 0.205 & 0.370 & 0.205 & 0.368 & 0.202 & \textbf{0.364} \\
    
    \midrule
    
    \multirow{5}*{ETTm$_1$}
    & 24 & 0.015 & 0.092 & 0.016 & 0.093 & \textbf{0.014} & 0.089 & \textbf{0.014} & \textbf{0.087} & 0.015 & 0.090\\
    & 48 & 0.028 & 0.124 & 0.034 & 0.137 & \textbf{0.027} & 0.123 & \textbf{0.027} & \textbf{0.121} & 0.028 & 0.125 \\
    & 96 & 0.044 & 0.159 & 0.049 & 0.169 & 0.045 & 0.162 & 0.044 & 0.158 & \textbf{0.043} & \textbf{0.156} \\
    & 288 & \textbf{0.086} & \textbf{0.225} & 0.099 & 0.252 & 0.097 & 0.238& 0.092 & 0.233 & 0.094 & 0.234 \\
    & 672 & \textbf{0.124} & \textbf{0.271} & 0.149 & 0.296 & 0.144 & 0.289& 0.137 & 0.289 & 0.134 & 0.280 \\

    \midrule
    
    \multirow{5}*{Weather}
    & 24 & \textbf{0.037} & \textbf{0.152} & 0.131 & 0.317 & 0.105 & 0.276& 0.049 & 0.198 & 0.080 & 0.257  \\
    & 48 & \textbf{0.024} & \textbf{0.114} & 0.094 & 0.226 & 0.077 & 0.202& 0.034 & 0.153 & 0.049 & 0.181 \\
    & 168 & \textbf{0.012} & 0.082 & 0.034 & 0.120 & 0.030 & 0.118& 0.013 & \textbf{0.080} & 0.017 & 0.086 \\
    & 336 & \textbf{0.009} & \textbf{0.069} & 0.021 & 0.097 & 0.019 & 0.094 & 0.011 & 0.074 & 0.012 & 0.074 \\
    & 720 & 0.011 & 0.083 & 0.015 & 0.086 & 0.014 & 0.084 & \textbf{0.009} & \textbf{0.070} & 0.011 & 0.075 \\

    \midrule
    
    \multicolumn{2}{l}{Avg.} & \textbf{0.088} & \textbf{0.211} & 0.103 & 0.235  & 0.099 & 0.229 & 0.092 & 0.217 & 0.096 & 0.224 \\
    
    \bottomrule
  \end{tabular}
  }
  
  \label{forecast-univar}
  % \vskip -0.1in
\end{table}

\subsection{Ablation study}

To assess the contribution of each component in our proposed method, we conduct an ablation study by systematically modifying key elements and evaluating the resulting performance. Specifically, we analyze the effect of single versus multiple positives with and without negatives. The results, presented in Table \ref{tab:ablation}, demonstrate that removing multiple positives leads to a drop in performance and training stability, highlighting their importance in learning useful representations. Similarly, not using the entire batch as negatives degrades the model’s ability to capture meaningful temporal dependencies. These findings validate the effectiveness of our design choices and underscore the necessity of each component in enhancing both representation quality and downstream generalization.

The ablation study also compares the performance of sequential negative sampling versus shuffled negative sampling within the MP-Xent framework. Interestingly, the results show that both strategies lead to comparable performance across both forecasting and classification tasks. While shuffled negative sampling does provide a marginal improvement in the \textsc{SleepEeg} data, it comes at the cost of additional computational and memory requirements. Specifically, shuffling the negative samples requires maintaining an extra similarity matrix for the shuffled sequence, which increases both memory usage and computational overhead.

\begin{table}[ht]
\centering
\caption{Ablation study on the effect of single versus multiple positives with and without negatives. Results are reported for long-term forecasting (MSE) and classification (accuracy). The 'Multiple positive + negatives (MP-Xent)' configuration achieves the best performance, showing significant improvements in both tasks, with percentage changes relative to the 'Single positive w/o negatives' baseline. All classification tasks are averaged over seeds 1-5 for reproducibility. Entries marked with '-' indicate that NaN values were encountered, underscoring the training instability when using single positives.}
\vskip 0.15in
\resizebox{0.48\textwidth}{!}{ 
  \begin{tabular}{llllllll}
    \toprule
    \multirow{3}{*}{Model Variants} & \multicolumn{4}{c}{\textit{forecasting}}  \\
    \cmidrule(lr){2-5} % Half-rule lines for precision and recall columns
    &  ETTh$_1$ & ETTh$_2$ & ETTm$_1$ & Weather \\
    \midrule

    Single positive w/o negatives & 0.210 & 0.211 & 0.130 & 0.014\\
    
    Single positive + negatives & - & - & 0.121 & 0.012 \\

    \midrule
    Multiple positive + shuffled negatives (MP-Xent) & 0.189 & 0.201 & 0.119 & 0.011\\
    
    Multiple positive + negatives (MP-Xent) & 0.189 (-10.0\%)  & 0.201 (-4.7\%) & 0.124 (-4.6\%)  & 0.011 (21.4\%)\\

    \midrule

        \\
    
    % \midrule
    & \multicolumn{3}{c}{\textit{classification}}  \\
    \cmidrule(lr){2-4} % Half-rule lines for precision and recall columns
    & \textsc{Harth} & \textsc{SleepEeg} & \textsc{Ecg}\\
    \midrule

    Single positive w/o negatives & 82.92 & 55.53 & 63.92\\
    
    Single positive + negatives & 90.30 & 61.91 & 73.16 \\
    \midrule
    Multiple positive + shuffled negatives (MP-Xent) & 90.01 & 62.97 & 76.61\\
    
    Multiple positive + negatives (MP-Xent) & 90.33 (+8.9\%) & 62.34 (+12.3\%) & 79.06 (+23.7)\\

    \bottomrule
  \end{tabular}
}
\vskip -0.1in
\label{tab:ablation}
\end{table}

We further investigate the impact of two key hyperparameters—temperature ($\tau$) and batch size—on model performance, with results summarized in Table \ref{tab:ablation2}. For forecasting tasks, the model exhibits robust stability across different settings: varying $\tau$ from 0.1 to 1.0 and adjusting batch size from 16 to 64 results in negligible changes in MSE, particularly for ETTh$_1$, ETTh$_2$, and Weather datasets.

In classification, the effects of these hyperparameters are slightly more pronounced but remain modest overall. For temperature, a lower $\tau = 0.1$ generally yields better accuracy on SleepEeg and Harth, while $\tau = 0.5$ and $\tau = 1.0$ favor ECG. Batch size shows similar trends as performance remains consistent across most datasets, though larger batches (e.g., 64) lead to a slight drop in accuracy on Harth and ECG. These findings suggest that while our model is not highly sensitive to temperature or batch size, tuning $\tau$ and batch size may yield marginal gains, especially for certain classification tasks.

\begin{table}[ht]
\centering
\caption{Effect of varying batch size and temperature on model performance. Results are reported for long-horizon forecasting (MSE) and classification (accuracy). For reproducibility, all classification results are averaged over seeds 1–5.}
\vskip 0.15in
\resizebox{0.3\textwidth}{!}{ 
  \begin{tabular}{llllllll}
    \toprule
    & \multicolumn{3}{c}{\textit{forecasting}}  \\
    \cmidrule(lr){2-5} % Half-rule lines for precision and recall columns
    &  ETTh$_1$ & ETTh$_2$ & ETTm$_1$ & Weather \\
    \midrule

    \texttt{$\tau$ = 0.1} & 0.189 & 0.201 & 0.125 & 0.011\\
    
    \texttt{$\tau$ = 0.5} & 0.189 & 0.201 & 0.124 & 0.011 \\
    \texttt{$\tau$ = 1.0} & 0.189 & 0.201 & 0.163 & 0.012\\

    \midrule
    \midrule

    \texttt{Batch = 16} & 0.189 & 0.201 & 0.134 & 0.008\\
    
    \texttt{Batch = 32} & 0.189 & 0.201 & 0.134 & 0.008 \\
    \texttt{Batch = 64} & 0.189 & 0.201 & 0.134 & 0.008\\
    
    \midrule

    % \midrule
     & \multicolumn{3}{c}{\textit{classification}}  \\
    \cmidrule(lr){2-4} % Half-rule lines for precision and recall columns
    & \textsc{Harth} & \textsc{SleepEeg} & \textsc{Ecg}\\
    \midrule

    \texttt{$\tau$ = 0.1} & 90.13 & 62.34 & 69.43 \\
    
    \texttt{$\tau$ = 0.5} & 89.64 & 59.43 & 79.06  \\
    \texttt{$\tau$ = 1.0} & 89.41 & 60.47 & 78.86 \\

    \midrule
    \midrule

    \texttt{Batch = 16} & 90.15 & 62.58 & 74.08 \\
    
    \texttt{Batch = 32} & 90.07 & 62.90 & 66.33  \\
    \texttt{Batch = 64} & 88.21 & 63.35 & 68.96 \\

    \bottomrule
  \end{tabular}
}
\vskip -0.1in
\label{tab:ablation2}
\end{table}

\subsection{Anomaly Detection}

In addition to TS classification and forecasting, we also evaluate CaTT on anomaly detection against TS representation learning methods: TS2Vec, CoST, TS2Vec + \textit{SoftCLT}, and MF-CLR, across nine multivariate anomaly detection datasets from TSB-AD benchmark datasets \cite{tsbad}, namely Daphnet, GECCO, Genesis, LTDB, MSL, SMAP, SMD, SWaT and TOA. Full dataset descriptions are provided in the appendix. 

In an anomaly detection setup, methods are trained on normal instances and aim to predict anomalies in the test set. We want to investigate whether CaTT's ability to bring similar instances closer during training will help distinguish normal instances from anomalous ones. We train the feature extractor of all baselines on the normal dataset in an unsupervised manner. During inference, scores are computed on the learned features as the sum of the projected distances of a sample on all eigenvectors, as described in \citet{shyu2003novel} and \citet{pca}. The anomaly score is calculated as the sum of the weighted Euclidean distances from the sample to the hyperplane formed by the selected eigenvectors. We train on each dataset for 15 epochs, using the same hyperparameter settings and model architecture as in the classification task and evaluated using the same protocol in \cite{tsbad}.

\begin{figure}[ht!]
\hspace*{1cm}  % Adjust the value to move the figure left
\includegraphics[scale=0.4]{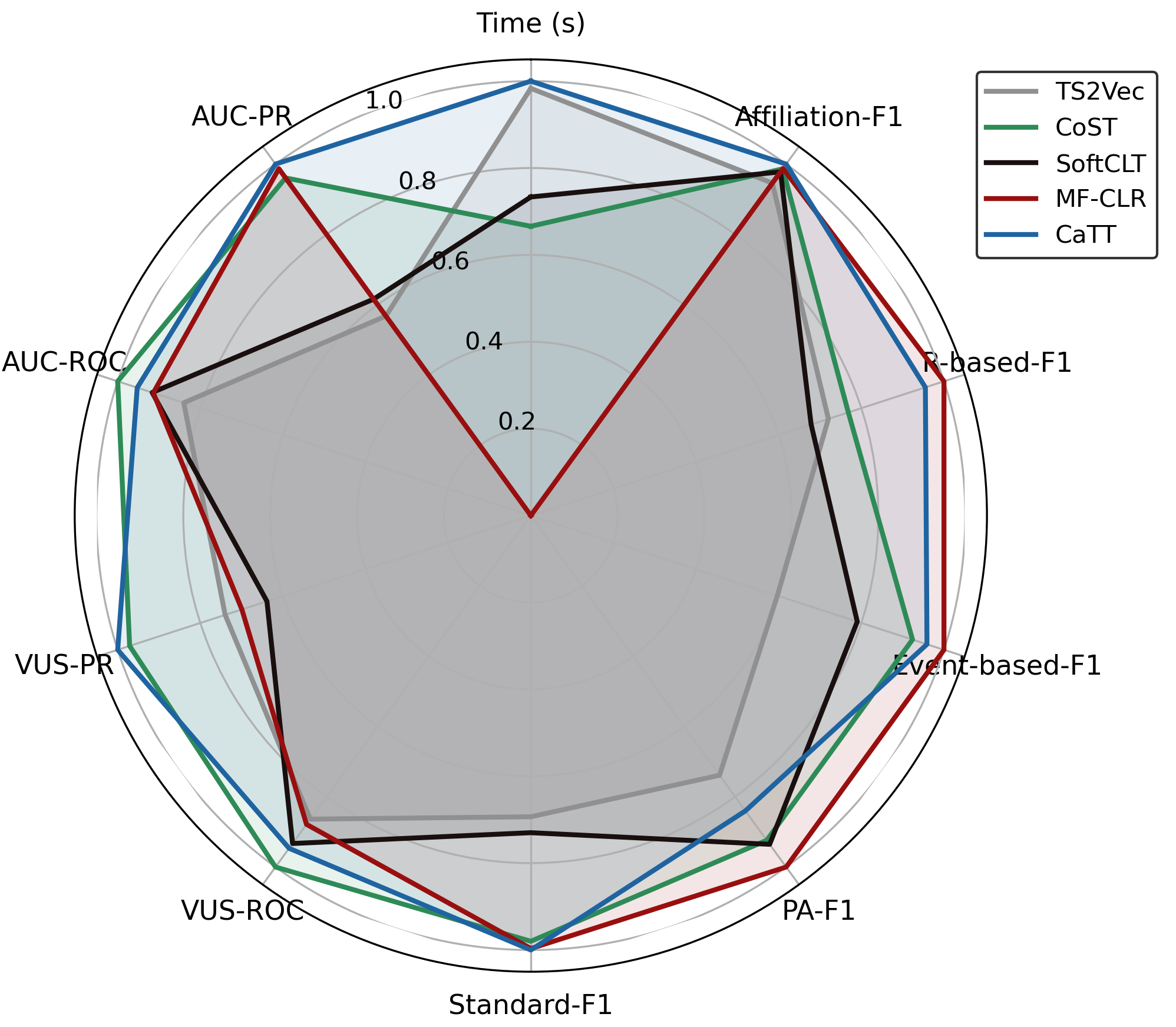}
\caption{Radar chart comparing the performance of CaTT with baseline methods across multiple anomaly detection metrics. Results are averaged across nine multivariate anomaly detection datasets (full details in the appendix). Time is log-normalized and inverted, so better (faster) methods appear outward on the plot.}
\label{fig:anomaly}
\end{figure}

The radar chart in Figure~\ref{fig:anomaly} shows performance across ten key metrics, including detection precision (\textit{AUC-PR}, \textit{VUS-PR}), robustness (\textit{R-based-F1}, \textit{Affiliation-F1}), and efficiency (\textit{Time}). All results are averaged across datasets, with time log-normalized and inverted so that faster methods appear outward on the plot. CaTT performs well in \textit{VUS-PR}, a recently proposed metric that integrates the entire Precision-Recall surface \cite{vuspr}. \textit{VUS-PR} has been found to be a reliable and consistent evaluation measure \citep{tsbad}. On this metric, CaTT shows competitive performance compared to other methods, highlighting the potential for learning representations that are useful for anomaly detection. Additionally, CaTT is the fastest method, making it suitable for real-world, large-scale deployments.

\section{Conclusion}

In this work, we present CaTT, a method for unsupervised representation learning of time series data. The CaTT method demonstrates the ability to learn semantically meaningful representations off the shelf and outperforms previous time series representation learning methods in downstream classification and forecasting tasks. Additionally, we show that CaTT models not only learn useful representation at the instance level but is also capable of learning timestamp-level representation, as underscored by its performance in time series forecasting. Finally, we studied the contribution of individual components of CaTT. We conclude that CaTT is the best performing model with multiple positives. This shows that using a multiple-positive sampling strategy, where adjacent time steps are selected as positives in an NT-tuple loss, allows our model to compete with previous approaches. These include methods that rely on statistical techniques, masking, and prediction sampling \citep{tnc, timedrl, simmtm}, as well as approaches that use augmentations \citep{infots, ts2vec, cost, soft, mfclr}. Also, the CaTT model not only delivers exceptional performance in downstream tasks but also exhibits the shortest training time (Table \ref{fig:pretime} and Table \ref{tab:linear}). This highlights the efficiency of the multiple positive sampling strategy in our MP-Xent contrastive objective, making it suitable for real-world applications, particularly in settings where inference speed is crucial.

\subsubsection*{Acknowledgments}

%\begin{ack}
 This publication was funded by SFI NorwAI (Centre for Research-based Innovation, 309834) and the Office of Naval Research. SFI NorwAI is financially supported by its partners and the Research Council of Norway. The views expressed in this article are those of the author(s) and do not reflect the official policy or position of the U.S. Naval Academy, Department of the Navy, the Department of Defense, or the U.S. Government.

 We also acknowledge the useful conversations and assistance of Dr. Frank Alexander Kraemer.
% \end{ack}

%%%%%%%%%%%%%%%%%%%%%%%%%%%%%%%%%%%%%%%%%%%%%%%%%%%%%%%%%%%%%%%%%%%%%%%%

%%% Use this command to include your bibliography file.
\bibliography{mybibfile}

\newpage
\appendix
\onecolumn

\section{Further Explanation of N-pair Loss}
\label{app:npair}

To select multiple positives, a naive approach involves looping through a batch $N$ and slicing temporal adjacent instances for all items in a batch, leading to a high time complexity. SoftCLT's \citep{soft} method selects adjacent instances by looping through the batch twice to precompute a matrix $N \times N$ with soft assignments, resulting in a similarly high time complexity. In contrast, our method efficiently combines positive selection and MP-Xent loss computation by adapting the N-pair loss approach introduced in \citep{npair} and \citep{simclr}. This adaptation allows each sample in a batch to contribute to an \(N+1\)-tuple loss.

\subsubsection*{Vanilla N-pair Loss (Single Positives):}
\begin{enumerate}
    \item Given a batch $N$ of time series instances, create an $N \times N$ similarity matrix by vector multiplication ($N \times N^T$).
    \item The lower diagonal elements represent positive pairs (numerator).
    \item The denominator is the sum of the column vectors minus the positives.
    \item Use the numerator and denominator to compute the $N$-tuple loss (Equation \ref{eqn:single}).
\end{enumerate}

\begin{figure*}[ht!]
\begin{center}
\includegraphics[scale=0.9]{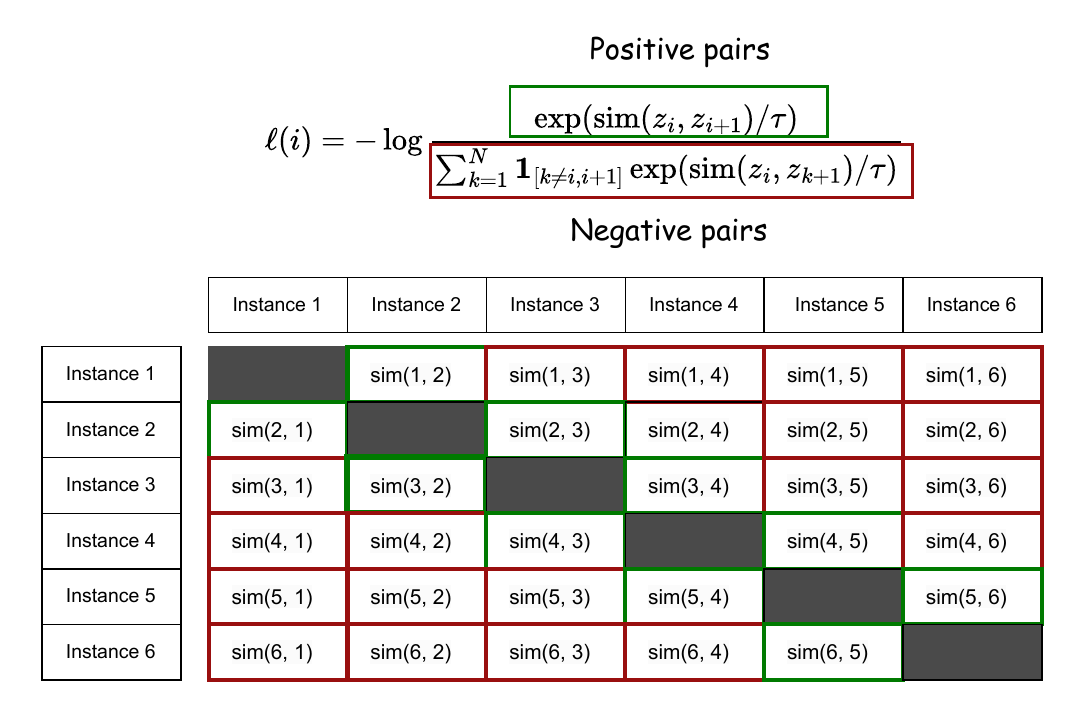}
\end{center}
\caption{Single positive derivation on 6 instances.}
\label{fig:sp}

\end{figure*}

\subsubsection*{N-pair Loss Extension (Multiple Positives):}
\begin{enumerate}
    \item Given a batch $N$ of time series instances, create an $NT \times NT$ similarity matrix by vector multiplication ($NT \times NT^T$).
    \item Select the lower diagonal elements.
    \item The positive pairs are the sum of the shifted left and right of the lower diagonal elements (numerator).
    \item The denominator is the sum of all elements in the similarity matrix (except the last two columns along each row).
    \item Subtract a combination of two lower diagonal slices from the sum to produce the denominator (Equation 3).
    \item Use the numerator and denominator to compute the $NT$-tuple loss (Equation 5).
\end{enumerate}

\begin{figure*}[ht!]
\begin{center}
\includegraphics[scale=0.9]{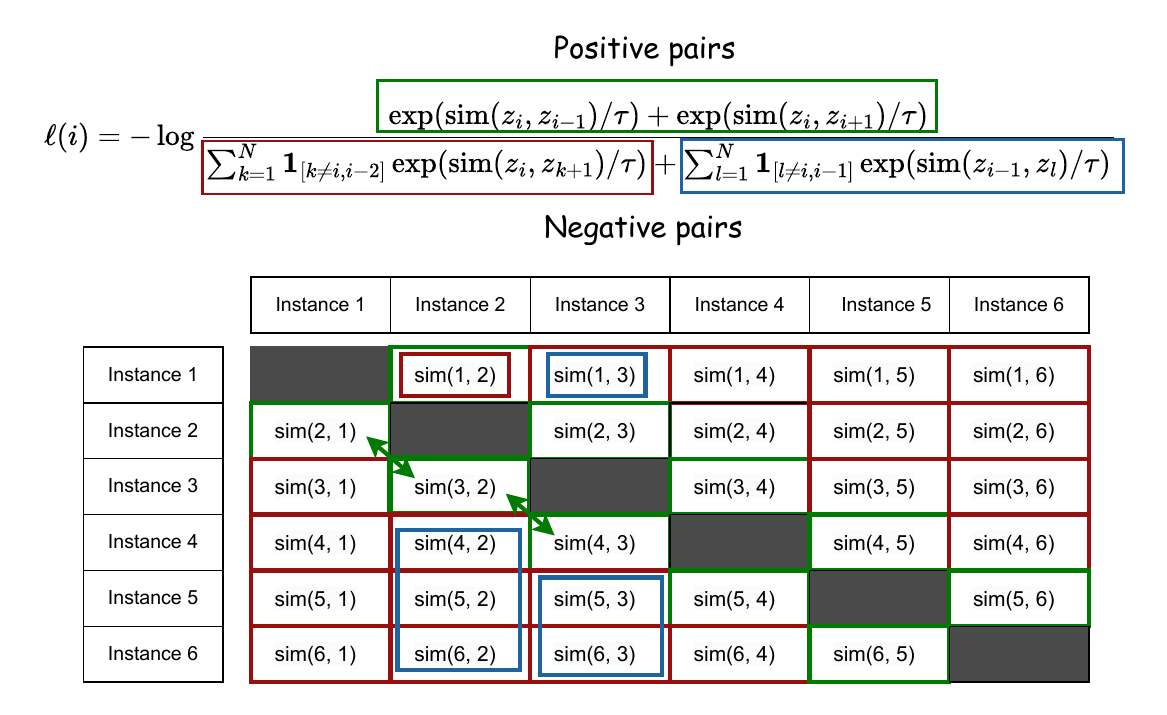}
\end{center}
\caption{Multiple positives derivation on 6 instances.}
\label{fig:mp}

\end{figure*}

\section{Full Description of Model Architecture}
\label{app:architecture}

For the classification tasks, the encoder model is designed to extract features from time-series data using 1D convolutional layers. It reduces the dimensionality of the input while retaining meaningful temporal information. The model consists of three 1D convolutional layers, each followed by batch normalization and ReLU activation to introduce non-linearity and stabilize the training process. The input to the model is a 3D tensor of shape (batch size, sequence length, input dimension), where the batch size is 8 for all datasets while the sequence length and dimension are dependent on the dataset. After passing through the three convolutional layers, the output dimensionality is reduced to embedding dimension = 32. Each convolutional layer applies a kernel size of 1 to focus on individual time steps, progressively reducing the number of channels from the input dimension to 128, 64, and 32. We apply batch normalization after each convolution to stabilize the activations, and ReLU activation functions introduce non-linearity, ensuring only positive values are passed through. The output tensor is reshaped back to the original order, returning a feature representation of shape (batch size, sequence length, embedding dimension). This architecture efficiently captures temporal dependencies while reducing the dimensionality, making it suitable for downstream tasks such as the classification and prediction of time-series data. For the forecasting task, we use the same dilated convolutional architecture introduced by \citet{ts2vec}

\section{Reproduction Details for Baselines for CaTT}
\label{app:pretrain}

In unsupervised representation learning, it is difficult to ascertain what hyperparameters will lead to superior performance in downstream tasks. One of the strengths of our method is that the MP-Xent loss used in CaTT has only 1 tunable hyperparameter, $\tau$. So, we use temperatures $\tau$ = 0.5, $\tau$ = 0.1, and $\tau$ = 0.05 for the \textsc{Ecg}, \textsc{SleepEeeg}, and \textsc{Harth} datasets respectively. For the time series forecasting task, however, we use $\tau$ = 0.5 all through. For the pretraining of all models, we maintain the same hyperparameters. Specifically, for the forecasting task, the batch size is set to 8, and the learning rate is 0.001. The number of optimization iterations is 200 for datasets smaller than 100,000 and 600 otherwise. The representation dimension is fixed at 320, following \citet{ts2vec}. We use the same setup for the classification tasks, except for setting the number of training iterations on the \textsc{Ecg} with a size of approximately 150k to 200 iterations. We train all models on an NVIDIA V100 GPU.

\section{Reproduction Details for Baselines}
\label{app:baseline}

In this section, we provide the reproduction details for the methods compared against. All results presented in this work are based on reproduction using code provided by the authors.

\textbf{InfoTS} \citep{infots}. We use the code and default parameters provided by the authors for the baseline. Specifically, we set the probabilities of the two different augmentation views as $p$ = 0.2, maximum train length = 500, and then the temperature used in contrastive loss functions $\tau_0$ and $\tau_1$ as 2.0 and 0.1, respectively. In the loss function, k=8 is used to define the number of local negatives for the local infoNCE loss function (again, default parameters by authors). Finally, we combine both the global and local infoNCE losses.

\textbf{TS2Vec} \citep{ts2vec}. We use the implementation and default parameters provided by the authors for the TS2Vec model. Specifically, we set the maximum sequence length during training to 500. The cropping is performed by selecting two random temporal windows within the sequence, defined by crop lengths and offsets dynamically generated during training. In each epoch, two augmented views of the input sequence are created: $x_1$ and $x_2$, where the lengths of the crops vary slightly. To ensure matching dimensions for the contrastive loss, padding is applied to equalize the output dimensions if one crop is shorter than the other. Finally, the hierarchical contrastive loss is computed based on these two views. We substituted the dilated CNN with our simple 1D CNN encoder to create a fair comparison across all baselines.

\textbf{TNC} \citep{tnc}. For the TNC, we adopt all relevant functions from the author code repository, namely: find neighbors, find non-neighbors, and binary cross entropy (BCE) loss function. The authors use a discriminator network to distinguish between two inputs, $x$ and $\bar{x}$, based on their similarity. The model architecture comprises two linear layers with a ReLU activation and dropout for regularization. Specifically, it concatenates the feature vectors of the two inputs into a single tensor, then fed through the model to output a probability score indicating whether the inputs belong to the same neighborhood. The weights of the linear layers are initialized using the Xavier uniform distribution. We use a Monte Carlo sample size and window size of 20, and w (hyperparameter to control the contribution of the different losses) as 0.1. All hyperparameters are used as provided by the authors and kept the same for all datasets.

\textbf{CoST} \citep{cost}. For this reproduction of the CoST baseline, we use the implementation and default parameters provided by the authors. The CoST method adapted the Dilated CNN from TS2Vec. To ensure all methods have the same backbone feature extractor, we replace this with our 1D CNN, which is used across all methods. The parameters used for this experiment are kernels = [1, 2, 4, 8, 16, 32, 64, 128], depth = 10, alpha = 0.05, K = 256, sigma = 0.5, and multiplier = 5.

\textbf{SimMTM} \citep{simmtm} is a pre-training framework designed for masked time-series modeling. This approach aims to preserve essential temporal variations that might be disrupted by random masking strategies. We use the official code by the authors for the reproductions. The core component of the method is the SimMTM loss. We use the temperature $\tau$ = 0.1, $\tau$ = 0.05, and $\tau$ = 0.5 for the SleepEEG, HARTH, and ECG datasets, respectively. We use the same temperature values for CaTT framework.

\textbf{TimeDRL} \citep{timedrl} focuses on disentangled representation learning for multivariate time-series data. It emphasizes capturing distinct factors of variation within the data, facilitating improved performance in downstream tasks such as forecasting and classification. We use the official parameters provided by the authors, such as patch length = 10, stride = 1, and enable channel independence = False.

\textbf{MF-CLR} \citep{mfclr} introduces a self-supervised framework designed to learn effective representations from multi-frequency time series. The method leverages contrastive learning while incorporating a hierarchical mechanism that spans across different frequencies along the feature dimension. We use the default parameters provided by the authors but changed to projection dimension to match other models to 320 (temporal unit=0, ph dim = 320, hidden dims=64, depth=10, projection= True, da= "proposed"). The authors provided hardcoded values for the grain split for the UEA datasets. Since we train on entirely different data sets, without intuition on how the author came up with these values, we replicated the dataset input dimension to form the grain split list.

\textbf{SoftCLT} \citep{soft}. SoftCLT aims to overcome the issue of ignoring inherent correlations between adjacent timestamps in a sequence. This method is not a standalone architecture and is built on existing contrastive learning approaches to enhance performance. According to their paper, the best-performing baseline is obtained by combining SOftCLT with TS2Vec. So, we use the author's implementation of this as a comparison in both the classification and forecasting tasks. The authors' best-performing model is the use of Dynamic Time Warping (DTW) in computing the soft assignments. Due to the time and computation complexity required in computing DTW for the real-world long sequences used in our work, we opted for the Cosine similarity distance provided by the authors as a substitute.

% \begin{table*}[ht]
% \centering
% \caption{Dataset distributions used across all experiments. We pre-train all models for 500 epochs on 80\% of the entire data instances and evaluate downstream performance on the remaining 20\%. For the \textsc{Harth} and \textsc{Ecg}, each instance is 1 second while for the \textsc{SleepEeg} an instance is 2 seconds.}

% \resizebox{0.8\textwidth}{!}{ 
%   \begin{tabular}{lccccc}
%     \toprule
%     & \# Train instances & \# Test instances & Dimensions & Classes \\
%     \midrule

%     \textbf{\textsc{Harth}} & 1,016,141 & 253,946 & 156 & 12  \\
%     \textbf{\textsc{SleepEeg}} & 296,700 & 74,100 & 178 & 5 \\
%     \textbf{\textsc{Ecg}} & 1,225,416 & 306,355 & 252 & 4 \\
    
%     \bottomrule
%   \end{tabular}
% }
% \label{tab:datadist}
% \end{table*}

\section{Details for Benchmark Tasks}

\subsection{Linear evaluation with frozen backbone} The classification datasets are preprocessed into small blocks of short instances. To evaluate the instance-level representations on time series classification, we train a linear model on the features from the frozen backbone for 10 epochs. The results presented in Figure \ref{tab:linear} are average on all seed values. 

\subsection{Semi-supervised classification} For the semi-supervised classification, we maintain the same setup as in the frozen backbone. The major difference is the selection of random 5\%, 10\%, and 10\% of the training dataset for fine-tuning and averaging over 15 runs (5 runs per model seed).

\subsection{Time series forecasting} To evaluate the timestamp-level
representations on time series forecasting, we follow the ridge regressing protocol in TS2Vec \citep{ts2vec}. We adopted the same evaluation setup and hyperparameters as used in TS2Vec. We presented the results for Mean Squared Error (MSE) and Mean Absolute Error (MAE) in Table \ref{forecast-univar}

%%%%%%%%%%%%%%%%%%%%%%%%%%%%%%%%%%%%%%%%%%%%%%%%%%%%%%%%%%%%%%%%%%%%%%%%%%%%%%%
%%%%%%%%%%%%%%%%%%%%%%%%%%%%%%%%%%%%%%%%%%%%%%%%%%%%%%%%%%%%%%%%%%%%%%%%%%%%%%%

\newpage
\section{Visualization of learned representations}

\label{app:tsne}

\begin{figure}[ht]
\begin{center}
\includegraphics[scale=0.09]{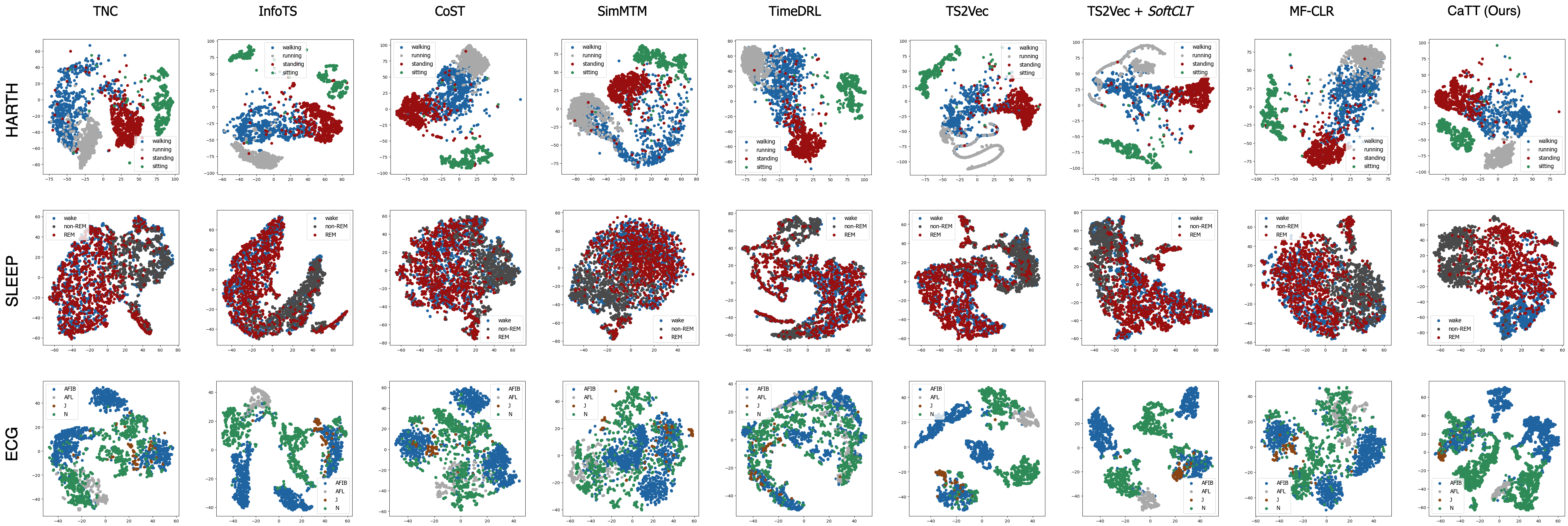}
\end{center}
\caption{t-SNE visualization of the learned embeddings on random instances on the \textsc{Harth} (first row), \textsc{Sleep} (second row), and \textsc{Ecg} (third row) test sets across all methods. For the \textsc{SleepEeg} each instance (data point) spans 2 seconds, while for the \textsc{Harth} and \textsc{Ecg} each instance is 1 and 6 seconds respectively.}
\label{fig:sleepfeauresappedix}

\end{figure}

\section{Long and short time series forecasting with TimeDRL baseline} 

TimeDRL \citep{timedrl}, which also tackles positive pair selection without explicit augmentations (using dropout instead) is another recent work with a focus on time series forecasting. The original work use a transformer architecture as the backbone feature extractor. We substituted the transformer backbone with the same dilated convolutional backbone as with the other baselines. We have presented the results in Table \ref{forecast-univar-ext}. The missing rows indicate datasets where TimeDRL returns NaN values (ETTm$_1$ and Weather).

\begin{table}[h]
  \centering
  \caption{Short-term and long-term forecasting results across multiple datasets (ETTh$_1$, ETTh$_2$, ETTm$_1$, Weather) and forecast horizons (H = 24, 48, 168, 336, 720). Performance is evaluated using Mean Squared Error (MSE) and Mean Absolute Error (MAE). CaTT demonstrates competitive accuracy compared to baseline methods, with the best results highlighted in bold.}
  \vskip 0.15in
  \scalebox{0.9}{
  \begin{tabular}{lccccccccccccccc}
  \toprule
         &  & \multicolumn{2}{c}{CaTT} &
         \multicolumn{2}{c}{TS2Vec} &
         \multicolumn{2}{c}{TS2Vec + \textit{SoftCLT}} &
         \multicolumn{2}{c}{CoST} &
         \multicolumn{2}{c}{MF-CLR} &
         \multicolumn{2}{c}{TimeDRL}\\
    \cmidrule(r){3-4} \cmidrule(r){5-6} \cmidrule(r){7-8} \cmidrule(r){9-10} \cmidrule(r){11-12} \cmidrule(r){13-14} \cmidrule(r){15-16} 
    Dataset & H & MSE & MAE & MSE & MAE & MSE & MAE & MSE & MAE & MSE & MAE & MSE & MAE\\
    \midrule
    \multirow{5}*{ETTh$_1$}
    & 24 & 0.040 & 0.152 & 0.040 & 0.153 & 0.040 & 0.153 & 0.041 & 0.152 & \textbf{0.039} & \textbf{0.149} & 0.047 & 0.166 \\
    
    & 48 & 0.063 & 0.193 & 0.063 & 0.193 & 0.063 & 0.192& 0.066 & 0.196 & \textbf{0.062} & \textbf{0.189} & 0.072 &0.207\\
    
    & 168 & \textbf{0.122} & \textbf{0.269} & 0.145 & 0.296 & 0.144 & 0.296 & 0.126 & 0.273 & 0.142 & 0.297&0.137&0.291 \\
    
    & 336 & \textbf{0.143} & \textbf{0.296} & 0.164 & 0.321 & 0.164 & 0.321& 0.150 & 0.305 & 0.165 & 0.323 &0.157&0.317\\
    
    & 720 & 0.189 & 0.354 & \textbf{0.174} & \textbf{0.340} & \textbf{0.174} & 0.342 & 0.188 & 0.354 & 0.201 & 0.371 &0.214&0.384\\
    
    \midrule
    
    \multirow{5}*{ETTh$_2$}
    & 24 & 0.095 & 0.235 & 0.090 & 0.227 & \textbf{0.089} & \textbf{0.226}& 0.091 & 0.231 & 0.093& 0.230  &0.091&0.232\\
    & 48 & 0.130 & 0.279 & 0.126 & 0.275 & \textbf{0.125} & \textbf{0.274}& 0.127 & 0.277 & 0.133 & 0.279 &0.128&0.278\\
    & 168 & \textbf{0.190} & \textbf{0.348} & 0.200 & 0.357 & 0.201 & 0.357& 0.206 & 0.359 & 0.200 & 0.352 &0.201&0.355\\
    & 336 & \textbf{0.196} & \textbf{0.357} & 0.205 & 0.365 & 0.206 & 0.366 & 0.206 & 0.363 & 0.206 & 0.363 &0.205&0.363\\
    & 720 & 0.201 & 0.365 & 0.205 & 0.370 & 0.205 & 0.370 & 0.205 & 0.368 & 0.202 & 0.364 &\textbf{0.195}&\textbf{0.359}\\
    
    \midrule
    
    \multirow{5}*{ETTm$_1$}
    & 24 & 0.015 & 0.092 & 0.016 & 0.093 & \textbf{0.014} & 0.089 & \textbf{0.014} & \textbf{0.087} & 0.015 & 0.090& - & -\\
    & 48 & 0.028 & 0.124 & 0.034 & 0.137 & \textbf{0.027} & 0.123 & \textbf{0.027} & \textbf{0.121} & 0.028 & 0.125& - & - \\
    & 96 & 0.044 & 0.159 & 0.049 & 0.169 & 0.045 & 0.162 & 0.044 & 0.158 & \textbf{0.043} & \textbf{0.156} & - & -\\
    & 288 & \textbf{0.086} & \textbf{0.225} & 0.099 & 0.252 & 0.097 & 0.238& 0.092 & 0.233 & 0.094 & 0.234 & - & -\\
    & 672 & \textbf{0.124} & \textbf{0.271} & 0.149 & 0.296 & 0.144 & 0.289& 0.137 & 0.289 & 0.134 & 0.280 & - & -\\

    \midrule
    
    \multirow{5}*{Weather}
    & 24 & \textbf{0.037} & \textbf{0.152} & 0.131 & 0.317 & 0.105 & 0.276& 0.049 & 0.198 & 0.080 & 0.257 & - & - & \\
    & 48 & \textbf{0.024} & \textbf{0.114} & 0.094 & 0.226 & 0.077 & 0.202& 0.034 & 0.153 & 0.049 & 0.181 & - & -\\
    & 168 & \textbf{0.012} & 0.082 & 0.034 & 0.120 & 0.030 & 0.118& 0.013 & \textbf{0.080} & 0.017 & 0.086 & - & -\\
    & 336 & \textbf{0.009} & \textbf{0.069} & 0.021 & 0.097 & 0.019 & 0.094 & 0.011 & 0.074 & 0.012 & 0.074 & - & -\\
    & 720 & 0.011 & 0.083 & 0.015 & 0.086 & 0.014 & 0.084 & \textbf{0.009} & \textbf{0.070} & 0.011 & 0.075 & - & -\\

    \midrule
    
    \multicolumn{2}{l}{Avg.} & \textbf{0.088} & \textbf{0.211} & 0.103 & 0.235  & 0.099 & 0.229 & 0.092 & 0.217 & 0.096 & 0.224 & 0.145 &0.295\\
    
    \bottomrule
  \end{tabular}
     }
  \label{forecast-univar-ext}
 
\end{table}

\begin{table}[ht]
\centering
\caption{Unsupervised pretraining time (in seconds) for short-term and long-term forecasting.}
\vskip 0.15in

\resizebox{0.9\textwidth}{!}{ 
  \begin{tabular}{lcccccc}
    \toprule
    & \multicolumn{5}{c}{\textsc{Time (s)}} \\
    \cmidrule(lr){2-6} 
    
    & CaTT & TS2Vec & TS2Vec + \textit{SoftCLT} & CoST & MF-CLR  & TimeDRL\\
    \midrule

    ETTh1 & \textbf{4.62} & 16.12 & 16.93 & 6.70 & 239.33 & 8.47\\
    ETTh2 & \textbf{3.02} & 15.63 & 18.50 & 6.69 & 216.32 & 4.89\\
    ETTm1 & \textbf{14.43} & 65.37 & 66.11 & 15.80 & 2604.35 & - \\
    Weather & 18.05 & 63.95 & 68.12 & \textbf{14.89} & 2332.83 &-\\
   \midrule
    Total & \textbf{40.12} & 161.07 & 169.66 & 44.08 & 5392.83 & -\\

    \bottomrule
  \end{tabular}
}
\label{tab:fore_time_2}

\end{table}

\newpage

\section{Anomaly Detection} 

\begin{table}[h!]
\centering
\caption{Multivariate time series anomaly detection dataset from \cite{tsbad}.}
\vskip 0.15in
\resizebox{0.95\textwidth}{!}{
\begin{tabular}{lllllllllll}
\toprule
\textbf{Name} & \textbf{\# TS Collected} & \textbf{\# TS Curated} & \textbf{Avg Dim} & \textbf{Avg TS Len} & \textbf{Avg \# Anomaly} & \textbf{Avg Anomaly Len} & \textbf{Anomaly Ratio} & \textbf{Category} \\
\midrule

Daphnet \cite{daph} & 17 & 1 & 9 & 38774.0 & 6.0 & 384.3 & 5.9\% & Seq \\

Genesis \cite{gene} & 1 & 1 & 18 & 16220.0 & 1.6 & 1267.0 & 3.8\% & Seq \\

SMD \cite{smd} & 28 & 22 & 38 & 25466.4 & 8.9 & 112.8 & 4.8\% & Seq \\
SWaT \cite{swat} & 1 & 1 & 59 & 207457.5 & 16.5 & 1093.6 & 12.7\% & Seq \\

SMAP \cite{smapmsl} & 54 & 27 & 25 & 7855.9 & 1.2 & 196.3 & 2.9\% & Seq \\
MSL \cite{smapmsl} & 27 & 16 & 55 & 3119.4 & 1.3 & 111.7 & 5.1\% & Seq \\

GECCO \cite{gecco} & 1 & 1 & 9 & 138521.0 & 51.0 & 33.8 & 1.2\% & Seq \\

LTDB \cite{ltdb} & 7 & 5 & 2 & 100000.0 & 105.0 & 134.4 & 15.5\% & Seq \\

TAO \cite{tao} & 45 & 13 & 3 & 10000.0 & 788.2 & 1.1 & 8.7\% & P\&Seq \\
\bottomrule
\end{tabular}
}
\vskip -0.1in
\label{tab:tsb_ad_m}
\end{table}

\end{document}